\documentclass[lettersize,journal]{IEEEtran}
\usepackage{amsmath,amsfonts}
\usepackage{algorithmic}
\usepackage{algorithm}
\usepackage{array}
\usepackage{subfig}
\usepackage{textcomp}
\usepackage{stfloats}
\usepackage{url}
\usepackage{verbatim}
\usepackage{graphicx}
\usepackage{cite}
\usepackage{tikz}
\usepackage{xcolor}
\usepackage{bm}
\usepackage{makecell}
\usepackage{tabu}
\usepackage{multirow}
\usepackage{enumitem}
\usepackage{colortbl}
\usepackage{multicol}
\usepackage{xspace}
\usepackage{bbding}
\usepackage{pifont}
\usepackage{tabularx}
\usepackage{array}    % 引入array包以支持新的列定义
\usepackage{hyperref} 
\usepackage{tcolorbox}
\newtheorem{theorem}{Theorem}
\newtheorem{proof}{{Proof}}
\newtheorem{definition}{{Definition}}

\newtheorem{problem}{{Problem}}

\definecolor{lightpurple}{RGB}{230, 190, 255}
\arrayrulecolor{black}

\begin{document}

\title{UniDyG: A Unified and Effective Representation Learning Approach for Large Dynamic Graphs}

% Bridging Continuous and Discrete Dynamic Graph Representation: An Effective and Robust Framework
% Bridging Continuous and Discrete Time Dynamic Graph Representation: A Fourier-Induced Solution
% Continuous-Time and Discrete-Time Dynamic Graph Representation: A Unified and Effective Framework

\author{Yuanyuan Xu,
Wenjie Zhang, Xuemin Lin~\IEEEmembership{Fellow,~IEEE}, and Ying Zhang
\thanks{Manuscript received November 2024. 
Corresponding author: Ying Zhang (email: ying.zhang@zjgsu.edu.cn).}
\thanks{Yuanyuan Xu and Wenjie Zhang are with the School of Computer Science and Engineering, The University of New South Wales, Sydney, NSW 2052, Australia~(e-mail: yuanyuan.xu@unsw.edu.au; wenjie.zhang@unsw.edu.au).

Xuemin Lin is with Antai College of Economics and Management, Shanghai Jiao Tong University, Shanghai 200052, china~(e-mail: xuemin.lin@gmail.com).

Ying Zhang is with the School of Statistics and Mathematics, School of Computer Science, Zhejiang Gongshang University, Hangzhou, Zhejiang 310018, China~(e-mail: ying.zhang@zjgsu.edu.cn).
}}
% ~\IEEEmembership{Graduate Member,~IEEE,}

\markboth{IEEE TRANSACTIONS ON KNOWLEDGE AND DATA ENGINEERING,~Vol., No., 2024}%
{Shell \MakeLowercase{\textit{et al.}}: A Sample Article Using IEEEtran.cls for IEEE Journals}


\maketitle

\begin{abstract}
Dynamic graphs, which capture time-evolving edges between nodes, are formulated in continuous-time or discrete-time dynamic graphs. They differ in temporal granularity: Continuous-Time Dynamic Graphs (CTDGs) exhibit rapid, localized changes, while Discrete-Time Dynamic Graphs (DTDGs) show gradual, global updates. This difference leads to isolated developments in representation learning for each type. To advance dynamic graph representation learning, recent research attempts to design a unified model capable of handling both CTDGs and DTDGs, achieving promising results. However, it typically focuses on local dynamic propagation for temporal structure learning in the time domain, failing to accurately capture the underlying structural evolution associated with each temporal granularity and thus compromising model effectiveness. In addition, existing works-whether specific or unified-often overlook the issue of temporal noise, compromising the model's robustness. To better model both types of dynamic graphs, we propose UniDyG, a unified and effective representation learning approach, which can scale to large dynamic graphs. Specifically, we first propose a novel Fourier Graph Attention (FGAT) mechanism that can model local and global structural correlations based on recent neighbors and complex-number selective aggregation, while theoretically ensuring consistent representations of dynamic graphs over time. Based on approximation theory, we demonstrate that FGAT is well-suited to capture the underlying structures in both CTDGs and DTDGs. We further enhance FGAT to resist temporal noise by designing an energy-gated unit, which adaptively filters out high-frequency noise according to the energy. Last, we leverage our proposed FGAT mechanisms for temporal structure learning and employ the frequency-enhanced linear function for node-level dynamic updates, facilitating the generation of high-quality temporal embeddings. Extensive experiments show that our UniDyG achieves an average improvement of $14.4\%$ over sixteen baselines across nine dynamic graphs while exhibiting superior robustness in noisy scenarios.

% demonstrate the effectiveness and robustness of our proposed UniDyG compared to thirteen baselines.

% In this paper, we aim to design a unified framework for representing both discrete-time and continuous-time dynamic graphs. We make critical observations for model design: (1) snapshot-level dynamics and edge-level dynamics; (2) static topology and temporal topology; and (3) model robustness. To achieve this, we propose UniDyG, a unified dynamic graph learning approach for both discrete-time and continuous-time dynamic graphs. To model both static and temporal topology in dynamic graphs, we explore a deeper and more nuanced structure learning in the frequency domain and propose a novel Fourier Graph Attention (FGAT), together with the decoupled framework, for structure learning. Furthermore, we design an energy-based To model temporal dynamics, UniDyG 
\end{abstract}


\section{Introduction}
\IEEEPARstart{D}{ynamic} graphs serve as a crucial data modality for representing time-evolving relationships (edges) between entities (nodes). They elegantly model a broad spectrum of real-world scenarios, including social networks~\cite{sun2022aligning}, e-commerce networks~\cite{liu2021anomaly}, chemical compounds~\cite{DBLP:journals/jmlr/KazemiGJKSFP20}, biological structures~\cite{DBLP:conf/bigdataconf/FuH22}, and traffic systems~\cite{DBLP:conf/kdd/KumarZL19}. According to temporal granularity, dynamic graphs can be classified into the Continuous-Time Dynamic Graph (CTDG) and the Discrete-Time Dynamic Graph (DTDG). The former views dynamic graphs as evolving \textit{events} (edges) in a streaming manner, preserving continuous dynamic and temporal structural changes. The latter treats dynamic graphs as a sequence of \textit{snapshots} that change at discrete intervals, storing static structures within each snapshot and temporal dynamics across snapshots.

% using static graph neural networks to capture static characteristics within each snapshot and temporal modeling to summarize historical information across snapshots. 

\noindent\textbf{Motivation.} Representation learning on dynamic graphs has emerged as a prominent research topic, which has been extensively studied for CTDGs~\cite{DBLP:conf/kdd/KumarZL19,DBLP:conf/iclr/TrivediFBZ19,tgn_icml_grl2020,zhou2022tgl,DBLP:journals/pacmmod/LiSCY23,gao2024simple,cheng2024co,TimeSGN} or DTDGs~\cite{DBLP:conf/icdm/SharanN08,DBLP:conf/www/ZhangYJL23,DBLP:conf/aaai/LiYZC0ZTWM23,zhu2023wingnn,DBLP:conf/kdd/YouDL22,hao2023dynamic,zhang2023dyted,zhao2024adversarial}, independently. Nevertheless, many real-world applications demand the simultaneous processing of both types, enabling more comprehensive analyses of complex, time-varying systems. For instance, e-commerce platforms need to predict quarterly purchasing trends while making real-time predictions for individual users. This dual requirement underscores the necessity for a unified approach that can seamlessly handle both discrete-time analysis (\textit{e.g.}, monthly trends) and continuous-time analysis (\textit{e.g.}, individual behaviors). Leveraging existing CTDG-specific or DTDG-specific methods for both graph types is technically feasible through graph conversion; however, this conversion brings significant drawbacks. Applying DTDG methods to CTDGs results in delayed predictions and information loss, whereas applying CTDG methods to DTDGs incurs information leakage and fails to capture global trends. Consequently, these methods easily yield sub-optimal performance and even limited utility for two types of dynamic graph learning.

Recent research attempts to develop a unified framework for both CTDGs and DTDGs, marking progress in dynamic graph learning. A notable effort in this line is the Decoupled dynamic Graph Neural Network (DGNN)~\cite{DBLP:journals/pvldb/ZhengWL23}, which separates temporal structure learning and dynamic learning into two stages. In the first stage, it designs a heuristic dynamic propagation algorithm to support local structure learning for both CTDGs and DTDGs. With the obtained structure embeddings, it subsequently trains different sequence models to capture temporal dynamics for various prediction tasks. However, the decoupled learning framework struggles to accurately capture complex graph evolution underlying dynamic graphs, which compromises model effectiveness. Additionally, it relies on clean data for temporal structure learning, making it difficult to generalize well in noisy learning scenarios. Motivated by the need for real-world applicability and to advance dynamic graph learning, our objective is to design a unified learning framework for both CTDGs and DTDGs that satisfies two key criteria: adaptable structure learning and strong robustness. We achieve this by addressing the following challenges.

 \begin{figure}[tbp]
    \centering
    \subfloat[Wikipedia (Time domain)]{\includegraphics[width=0.48\linewidth]{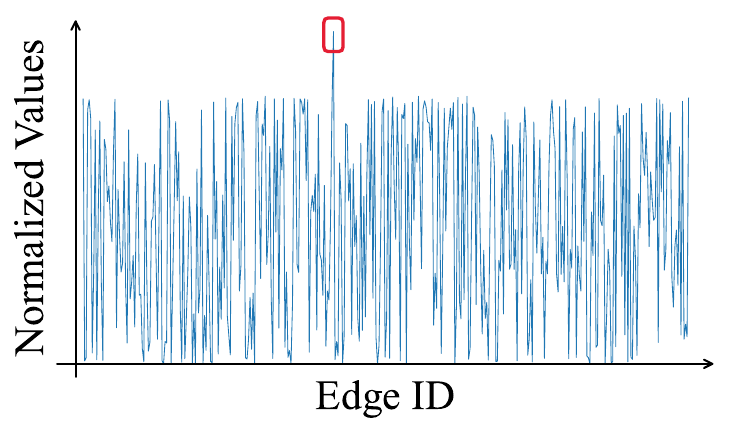}\label{fig:WIKI_T}}\;
    \subfloat[BitcoinAlpha (Time domain)]{\includegraphics[width=0.48\linewidth]{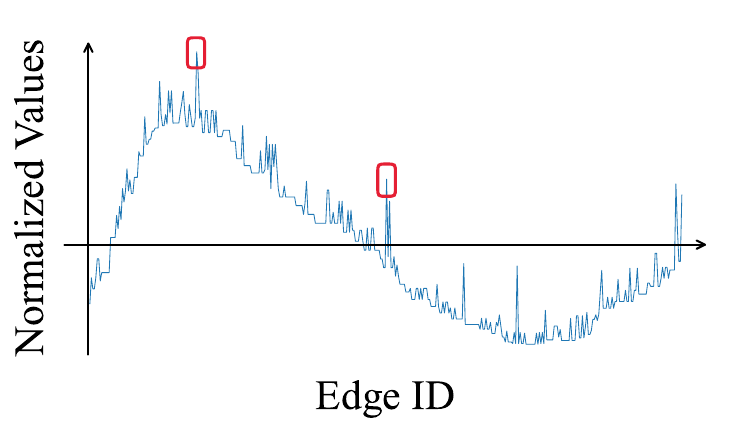}\label{fig:Alpha_T}}\\
    \subfloat[Wikipedia (Frequency domain)]{\includegraphics[width=0.48\linewidth]{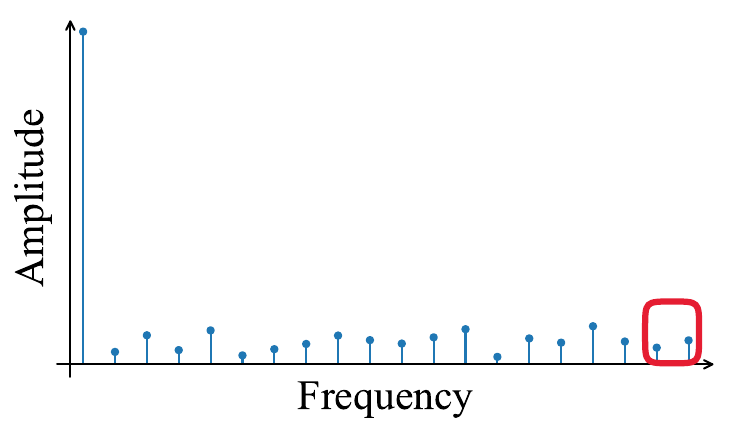}\label{fig:WIKI_F}} \;
    \subfloat[BitcoinAlpha (Frequency domain)]{\includegraphics[width=0.48\linewidth]{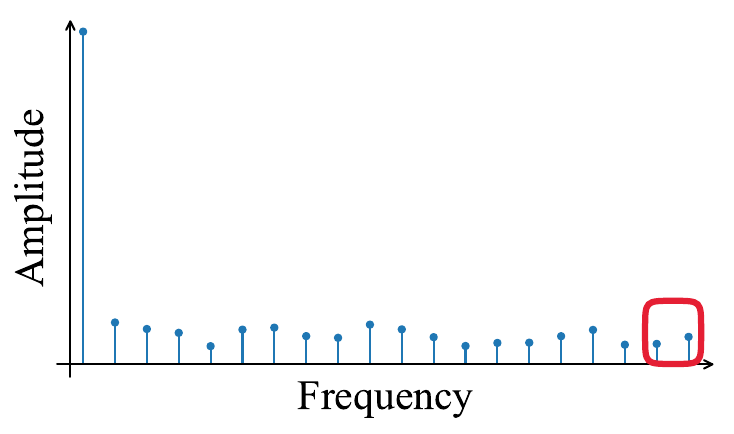}\label{fig:Alpha_F}}
    \caption{Visualization of selected edge attributes and timestamps in two domains on Wikipedia and BitcoinAlpha datasets.}
    \label{fig:edge_features}
\end{figure}

\noindent\textbf{Challenge~\uppercase\expandafter{\romannumeral1}: How to accurately model two types of dynamic graphs in a unified model?} According to the definition of dynamic graphs, CTDGs and DTDGs exhibit different graph evolution patterns. To illustrate this, we select a subgraph with $400$ edges from real-world datasets—Wikipedia (CTDG) and BitcoinAlpha (DTDG)—and visualize their graph signals in the time and frequency domains in Fig.~\ref{fig:edge_features}. In the time domain, the signals are significantly different: CTDGs display rapid, localized changes in Fig.~\ref{fig:WIKI_T}, while DTDGs show gradual, global trends in Fig.~\ref{fig:Alpha_T}. However, existing methods, whether unified or specific, typically focus on time-domain analysis and modeling, making it difficult to capture the distinct temporal dynamics and structures of both CTDGs and DTDGs. In contrast, looking at Figs.~\ref{fig:WIKI_F} and~\ref{fig:Alpha_F}, both types exhibit similar energy distributions concentrated in the low-frequency bands in the frequency domain. 
These insights motivate us to develop a frequency-domain approach that can more accurately capture the underlying patterns of both CTDGs and DTDGs.
\noindent\textbf{Challenge~\uppercase\expandafter{\romannumeral2}: How to efficiently capture global structural evolution?} Dynamic graphs often preserve long-term structural evolution, whether through continuous, real-time temporal events in CTDGs or discrete, static snapshots over time in DTDGs. This is reflected in the high amplitude of low-frequency signals, as shown in Figs.~\ref{fig:WIKI_F} and~\ref{fig:Alpha_F}. The existing unified approach typically prioritizes recent neighbors for dynamic propagation, thereby neglecting the broader historical context required to capture long-term structural evolution effectively. This easily leads to reduced performance across diverse dynamic graph types. Additionally, existing CTDG-specific and DTDG-specific methods also rely on local message passing and process graph structures within individual snapshots, respectively, which limits their receptive fields and thus compromises model effectiveness. Although increasing network depth can aid in capturing long-term structural evolution for specific approaches, it simultaneously incurs the neighbor explosion issue~\cite{DBLP:journals/pacmmod/LiSCY23,TimeSGN}, thereby resulting in prohibitive computation costs. Motivated by these, we aim to develop an efficient learning method capable of supporting global structural evolution from both CTDGs and DTDGs.

% A unified dynamic graph modeling approach must account for both types of temporal behaviors, capturing long-term structural correlations to identify consistent patterns in fast-evolving CTDGs and gradual trends in more stable DTDGs.

% Secondly, temporal noise, referred to as random or irrelevant short-term changes in graph structure or node/edge attributes, is ubiquitous in various real-world dynamic graphs due to measurement errors, temporary fluctuations, or random events not representative of underlying trends. For example, the red circle of Fig.~\ref{fig:C_E_T} could be noise information. Time-domain GNNs often lack inherent mechanisms to differentiate between meaningful changes and noise. Models may learn to fit these random fluctuations, mistaking noise for patterns. This leads to poor generalization and reduced model robustness. Time-domain GNNs are more susceptible to temporal noise and fluctuations.

\noindent\textbf{Challenge~\uppercase\expandafter{\romannumeral3}: How to enhance model robustness by mitigating the noise issue?} Temporal noise is ubiquitous in various real-world dynamic graphs, manifesting as random or irrelevant short-term fluctuations in graph structure or node/edge attributes. For instance, as shown in Figs.~\ref{fig:WIKI_T} and~\ref{fig:Alpha_T}, outliers (highlighted in red rectangles) are evident in selected subgraphs from the Wikipedia and BitcoinAlpha datasets, corresponding to high-frequency components in Figs.~\ref{fig:WIKI_F} and~\ref{fig:Alpha_F} (characterized by relatively low energy in red rectangles). However, most approaches—whether unified or specific—overlook the noise issue in dynamic graph learning. Consequently, they may learn to fit these random fluctuations, leading to reduced robustness. A recent CTDG-specific FreeDyG introduces a learnable frequency filter, which can filter out some noise. However, it does not adequately account for the characteristics of temporal noise in the frequency domain, potentially filtering out some valuable information along with the noise and thus compromising model effectiveness. We aim to accurately filter out temporal noise for better robustness.

\noindent\textbf{Our Contribution.} In this paper, we propose a UniDyG, a novel \underline{Uni}fied representation learning approach for both continuous-time and discrete-time \underline{Dy}namic \underline{G}raphs. We shift the focus from the time domain to the frequency domain for model design, while developing a unified data format and training pipeline to minimize information loss and information leakage, addressing \textbf{Challenge~\uppercase\expandafter{\romannumeral1}}. 
We then propose a novel Fourier Graph Attention (FGAT) mechanism to model underlying graph patterns by capturing local structural correlations from recent neighbors and global structural correlations through Fourier transform and complex-number selective aggregation. We theoretically demonstrate that FGAT possesses the Lipschitz continuity \textit{w.r.t.} time translations and the universal approximation property in the frequency domain, making it adaptable to temporal structures in both CTDGs and DTDGs and thus addressing \textbf{Challenge~\uppercase\expandafter{\romannumeral2}}. To tackle~\textbf{Challenge~\uppercase\expandafter{\romannumeral3}}, we optimize FGAT with an energy-gated unit that adaptively filters out high-frequency noise while preserving meaningful abrupt changes, dubbed FGAT\_N. Last, we leverage one-layer FGAT and FGAT\_N to implement a divided temporal message passing paradigm~\cite{TimeSGN}, allowing an effective and robust temporal structure learning; we employ a frequency-enhanced linear function with a global view to update node states, which can capture long-term temporal dynamics. In this way, our UniDyG can generate high-quality temporal embeddings while ensuring strong scalability and robustness. 

We summarize the main contributions as

\begin{itemize}
\item We propose a unified and effective representation learning approach for both continuous-time and discrete-time dynamic graphs, along with a unified training pipeline and data format designed to minimize information loss and prevent information leakage.

\item We propose a novel Fourier graph attention mechanism to capture underlying graph patterns by exploring both local and global structural correlations. We theoretically demonstrate that it can generate consistent and high-quality representations across diverse temporal structures over time.

% We prove that it is well-suited for unified temporal structure learning across both CTDGs and DTDGs. 
% while providing a theoretical guarantee for consistent representations of dynamic graphs over time.

\item UniDyG is designed to capture long-term temporal and structural dependencies over dynamic graphs while filtering out high-frequency temporal noise, thus enhancing the model's effectiveness and robustness.

% We prove that UniDyG is well-suited for unified learning across both CTDGs and DTDGs.

% while providing an optimized version for alleviating the temporal noise across diverse graph structures.

% We optimize our FGAT for mitigating the noise issue and implement them to generate temporal embeddings.

\item Extensive experiments validate that our UniDyG achieves an average improvement of $14.4\%$ across $9$ dynamic graphs compared to $16$ baselines. More importantly, it exhibits superior robustness under noise scenarios.
\end{itemize}

\section{Related Works}
We briefly review the research on dynamic graph representation learning and frequency-enhanced representation learning.

% However, the essence of both dynamic GNNs is to encode the temporal dynamics of the graph into updatable vector representations of nodes, which is also known as dynamic node embeddings.we first review the representation learning approaches over dynamic graphs from the perspectives of discrete-time, continuous-time, and both. Discrete-time dynamic graphs usually consist of discrete snapshots that reflect the periodic changes of the dynamic network; Continuous-time dynamic graphs keep all edges in one graph with a timestamp label, which can store the whole dynamic network efficiently and completely. Additionally, we briefly review the research on dynamic graph representation learning and frequency-enhanced representation learning.

\subsection{Representation Learning on Dynamic Graphs}  
Dynamic graph representation learning has been extensively studied, including three major research groups: CTDG-specific, DTDG-specific, and unified methods.

\noindent\textbf{CTDG-Specific Research Line.} Continuous-Time Dynamic Graphs (CTDGs) represent temporal interaction events occurring over continuous time, accommodating real-world scenarios with irregular temporal intervals. Temporal Graph Neural Networks (TGNNs) emerged as promising solutions for modeling complex dynamics and structures in CTDGs, focusing on three key aspects: generic frameworks, co-neighbor modeling, and computational efficiency. Specifically, a generic framework~\cite{tgn_icml_grl2020} of temporal graph neural networks for CTDGs was designed with static Graph Neural Networks (GNNs) for structure learning and Recurrent Neural Networks (RNNs) for temporal dynamics. Several works~\cite{DBLP:conf/kdd/KumarZL19,DBLP:conf/iclr/TrivediFBZ19,DBLP:conf/iclr/XuRKKA20,yang2021time}, including temporal graph attention networks~\cite{DBLP:conf/iclr/XuRKKA20}, can be viewed as specific instances of this framework, while numerous techniques~\cite{liu2020k,DBLP:conf/iclr/CongZKYWZTM23,DBLP:journals/pvldb/FangFGFH23,DBLP:conf/www/SureshSMN023,xu2024scalable,xu2024bootstrapping} were incorporated into this framework to enhance embedding quality. In addition to GNN-based structure learning, co-neighbor techniques~\cite{DBLP:conf/iclr/WangCLL021,souza2022provably,DBLP:conf/nips/JinLP22,luo2022neighborhood,yu2023towards,tian2023freedyg,li2024robust,cheng2024co} were developed to capture co-occurring neighbors underlying temporal graph patterns, thus improving prediction performance. However, these methods are often computationally intensive due to their complex learning mechanisms. To address this, various acceleration techniques were introduced to improve the scalability of TGNNs. For example, the learning process of TGNNs was decoupled based on different designs to enhance both time and memory efficiency~\cite{DBLP:conf/sigmod/WangLLXYWWCYSG21,TimeSGN}; or they optimize TGNNs by leveraging advanced techniques~\cite{zhou2022tgl,yang2023time,DBLP:journals/pvldb/LiSCY23,DBLP:journals/pacmmod/LiSCY23,gao2024simple}, such as cache management, temporal sampling, and data placement. These techniques enabled TGNNs to maintain accuracy while reducing computational overhead, improving their applicability in real-world scenarios. However, the above methods are exclusively tailored for CTDGs and are not easily adaptable to DTDG learning, which easily leads to the information leakage problem.

\noindent\textbf{DTDG-Specific Research Line.} Discrete-Time Dynamic Graphs (DTDGs) are represented through graph snapshots at different time intervals. DTDG-specific works fall into two main categories: matrix decomposition-based and GNN-based methods. Early studies~\cite{DBLP:conf/icdm/SharanN08,li2017attributed,DBLP:conf/icde/ZhuGYSG17,yang2021time} transformed dynamic graphs into static ones by aggregating and regularizing adjacency matrices of snapshots and then applied matrix decomposition techniques to generate node embeddings. However, these methods failed to capture the dynamic nature of graphs. Subsequently, GNN-based methods gained prominence, employing GNNs to model structure information within each snapshot while using various strategies for dynamic modeling across different snapshots. These methods include recurrent-based, gradient-based, hyperbolic-based, and disentangled-based methods. Recurrent-based methods~\cite{DBLP:conf/iconip/SeoDVB18,DBLP:conf/wsdm/SankarWGZY20,DBLP:conf/aaai/ParejaDCMSKKSL20,gao2022novel,DBLP:conf/www/BaiNZZY23,DBLP:conf/icde/0003SLRD23,DBLP:conf/kdd/0001ZKHK21,DBLP:conf/cikm/LiC21,DBLP:conf/www/ZhangYJL23,DBLP:conf/aaai/LiYZC0ZTWM23,qin2023high,zou2023event} used GNNs to generate node embeddings within each snapshot and leverage recurrent neural networks (\textit{e.g.}, Gated Recurrent Unit (GRU)) to model temporal dynamics across snapshots. While these methods are straightforward and intuitive, they increase the risk of overfitting due to excessive model parameters.
To enhance model robustness, gradient-based approaches~\cite{zhu2023wingnn,DBLP:conf/kdd/YouDL22,hao2023dynamic,zhao2024adversarial} leveraged gradient information to propagate snapshot dynamics and employed static GNNs for structure learning. Later, disentangled representation learning was applied to dynamic graph learning~\cite{zhang2023dyted,liu2021anomaly} to effectively identify different-level representations, achieving promising performance in downstream tasks. Additionally, to capture more complex structural information, such as hierarchical structures, some efforts were put into employing hyperbolic spaces to enhance the learning abilities of Euclidean GNNs for intricate structures~\cite{DBLP:conf/www/BaiNZZY23,DBLP:conf/kdd/0001ZKHK21}. However, DTDG-specific works generally fail to model global trends of the evolving topology between consecutive snapshots. In addition, if DTDG-specific methods are applied to CTDGs, this leads to information loss of fine-grained dynamics, thereby compromising model performance.

% Moreover, they typically require the aggregation of graph snapshots according to a predetermined time granularity, thus failing to accommodate real-world scenarios where interactions occur at irregular and diverse temporal intervals.

% Subsequently, path-based methods attempted to extract node/edge dynamic behaviors based on the graph structure and time constraints through various techniques, such as temporal random walks
% ~\cite{DBLP:conf/sigir/BianKDD19} and temporal point process~\cite{DBLP:conf/iclr/TrivediFBZ19,zuo2018embedding}

% Early methods adopt the matrix decomposition to capture the graph
% structure in each snapshot and regularize the smoothness of the
% representation of adjacent snapshots [1, 18]. Unfortunately, such
% matrix decomposition is usually computationally complex

%Attempts towards DTDG representation were done by employing matrix factorization techniques~\cite{du2023efficient}.

\noindent\textbf{Unified Research Line.} CTDGs and DTDGs often coexist in real-world applications, such as social networks. To accommodate these scenarios, researchers attempted to develop unified approaches that can handle both types of dynamic graphs, thereby advancing dynamic graph learning. Existing work first converted these dynamic graphs into a standardized input format by transforming CTDGs into discrete snapshots. Then, a unified decoupled model~\cite{DBLP:journals/pvldb/ZhengWL23} was proposed to process both CTDGs and DTDGs. It designed a heuristic dynamic graph propagation algorithm to generate structural embeddings and then leveraged sequence models to learn temporal dependencies for various downstream tasks within a two-step learning framework. However, such conversion into discrete snapshots directly leads to information loss, causing sub-optimal performance when modeling fine-grained dynamics in CTDGs. Furthermore, the distinct characteristics of CTDGs and DTDGs require different modeling priorities, but the existing unified work fails to satisfy the modeling requirements for both, \textit{i.e.}, complex and deep structure learning and global temporal evolution. In contrast, our unified approach comprehensively considers these characteristics across both CTDGs and DTDGs for temporal dynamic modeling and temporal structure learning, while avoiding information loss as well as improving model effectiveness and robustness.

% \subsection{Scalable Temporal Graph Neural Networks.} 
% Recently, researchers attempt to design scalable frameworks for training T-GNNs over dynamic graphs. Concretely, ROLAND~\cite{DBLP:conf/kdd/YouDL22} proposed an incremental training approach that did a truncated version of back-propagation-through-time, which significantly saved GPU memory cost for the DTDG training. TGL~\cite{zhou2022tgl} developed an efficient temporal graph sampling method and data parallel distributed training for large dynamic graph processing. Besides, the decoupled training method of static GNNs~\cite{DBLP:conf/kdd/BojchevskiKPKBR20,DBLP:conf/icml/WuSZFYW19} was also applied to T-GNNs~\cite{DBLP:journals/pvldb/ZhengWL23}. This T-GNN decoupled the message propagation and prediction processes during model training and can process both DTDGs and CTDGs. However, it also relied on static message passing for temporal structure learning. Different from these frameworks
% and approaches that focus on sampling and training optimization, TimeSGN takes an orthogonal approach by designing the temporal message passing paradigm and simple network to improve model scalability while maintaining expressiveness. 

\subsection{Frequency-enhanced Representation Learning}  
Fourier theory has been increasingly integrated into deep neural networks, achieving significant advancements in representation learning across various fields. In computer vision, neural networks were designed in the frequency domain, selectively preserving low-and high-frequency information for latent image restoration. This approach showed remarkable success in solving image deblurring challenges~\cite{kong2023efficient}. In Multivariate Time Series (MTS) forecasting, many models~\cite{zhang2017stock,yang2020adaptive,zhou2022fedformer,DBLP:conf/icml/Eldele0C0024} incorporated Fourier transform to capture complex periodic patterns in time series data, enhancing learning expressiveness and improving forecasting accuracy. In the context of dynamic graph learning, FreeDyG~\cite{tian2023freedyg} introduced a learnable filter in the frequency domain to help capture periodic temporal patterns and address the "shift" phenomenon in CTDGs. However, it still modeled both graph structures and temporal dependencies in the time domain. Moreover, it failed to explicitly filter out noise in the frequency domain, compromising model robustness to some extent. Given the inherent complexity of graph structures and temporal evolution, the full potential of frequency-domain methods in dynamic graph learning remains largely unexplored.

% Recently, disentangled representation learning has attracted a lot of research attention and achieved great success in many fields. Specifically, in computer vision, the identity of a
% face is disentangled from the views or pose information to perform better on image recognition. In natural language generation, the writing style is disentangled from the text content to serve
% the text-style transfer tasks. In graph neural networks, the factor behind the formation of each edge is disentangled for semisupervised node classification [41]. As demonstrated in existing
% research, the disentangling representation is an important step toward a better representation learning [20], which is much closer to human perception and cognition as well as can be more robust, explainable, and transferrable. 
% However, due to the complexity of graph structure and temporal evolution, how to learn disentangled representation in dynamic
% graphs remains largely unexplored.

\section{Preliminary and Background}
% In this paper, we aim to design a unified approach to handle two types of dynamic graphs, covering discrete-time and continuous-time dynamic graphs. For clarity, we first give their definitions and then present the problem definition. Additionally, we provide background knowledge behind our proposed approach.
\subsection{Dynamic Graphs}
\begin{definition} [Continuous-Time Dynamic Graph (CTDG)]\label{def:1}
A continuous-time dynamic graph $G = (U, E, X)$ is constructed by a sequence of temporal interaction events, where $U$ is the node set, $E$ is the edge set, and $X$ is the attribute matrix. $G = \{\delta(t_1), \delta(t_2),\cdots\}$ ordered by timestamps. Each temporal interaction event is a tuple $\delta(t) = (u_{i}, u_{j}, \bm{e}_{ij}(t), t)$ representing a temporal edge, where $u_{i}$ and $u_{j}$ are nodes, $\bm{e}_{ij}(t)$ denotes edge features, and $t$ is the timestamp at which $\delta(t)$ happens.
\end{definition}

Fig.~\ref{fig:overview}a shows a toy example of the CTDG. Here, a CTDG is a multigraph, implying that multiple interactions can occur between two nodes at different timestamps; for each interaction, edge features $\bm{e}_{ij}(t)$ may vary. 

\begin{definition} [Discrete-Time Dynamic Graph (DTDG)]\label{def:2}
A discrete-time dynamic graph $G=(U, E, X)$ is constructed by a sequence of graph snapshots, where $U$ is the node set, $E$ is the edge set, and $X$ is the attribute matrix. $G = \{G(t)\}_{t=1}^{T}$ ordered by timestamps. Each graph snapshot is a static graph $G(t) =\{ U(t), E(t), X(t)\}$ at timestamp $t\in [0, T]$, where $U(t), E(t), X(t)$ are the sets of nodes, edges, and attributes in $G(t)$.
\end{definition}

Fig.~\ref{fig:overview}a provides a toy example of the DTDG. Different from CTDGs, all edges within each snapshot in DTDGs have the same timestamp, which can be regarded as co-occurrence events. 

% That is, each edge can also be represented as a tuple $\delta(t) = (v_{i}, v_{j}, e_{ij}(t), t)$, where $v_{i}$ and $v_{j}$ are nodes, $e_{ij}(t)$ denotes edge features, and $t$ is the timestamp at which $G(t)$ happens.
%introduce their computation complexity O(|E|k^L) and delete irrelevant expression.

\subsection{Unified Format for CTDGs and DTDGs}\label{sec:3.2}
Definitions~\ref{def:1} and~\ref{def:2} describe two distinct data formats for dynamic graphs. To enable a unified modeling framework, it is essential to generate an input format that accommodates both CTDGs and DTDGs. Two potential strategies exist to achieve this: (1) Discretization of CTDG: converting a sequence of temporal events into a sequence of snapshots based on predefined time intervals (\textit{e.g.}, weekly); (2) Event-based format of DTDG: transforming each snapshot into a sequence of co-occurring events with the same timestamp. 

Prior work, such as DGNN~\cite{DBLP:journals/pvldb/ZhengWL23}, adopts the first approach by discretizing CTDG into snapshots based on time intervals. However, this leads to information loss within each snapshot, as each snapshot is treated as a static graph. In this paper, we opt for the second strategy, converting a DTDG into an event-based format. Concretely, for each edge $e_{ij}$ in the snapshot $G(t)$, we generate a corresponding event $\delta(t) = (u_{i}, u_{j}, \bm{e}_{ij}(t), t)$. This unified format preserves temporal granularity and minimizes information loss. However, it easily incurs the information leakage issue for training DTDGs, which will be addressed and discussed in Section~\ref{sec:4.4}. 
% , providing a solid foundation for developing a model capable of handling both continuous-time and discrete-time dynamic graphs effectively.

\subsection{Unified Problem Definition}
Building upon the unified input format introduced in Section~\ref{sec:3.2}, we aim to develop a unified representation learning approach for both continuous-time and discrete-time dynamic graphs. We formulate this problem within the \textit{encoder-decoder} paradigm~\cite{DBLP:journals/debu/HamiltonYL17}, focusing on the encoder function for dynamic graph representation learning.

\begin{problem}[Representation Learning on Dynamic Graphs]\label{problem:1}
Given a dynamic graph $G$ with a sequence of temporal edges $\delta(t)$, our aim is to design an encoder function $F: \delta(t)\rightarrow \bm{Z}_i(t), \bm{Z}_j(t)\in \mathbb{R}^d$, where $\bm{Z}_i(t), \bm{Z}_j(t)$ represent temporal embeddings of nodes $u_i$ and $u_j$ at time $t$, respectively.
\end{problem}

These temporal embeddings serve as the input to various decoder functions for specific downstream tasks, such as link prediction~\cite{tgn_icml_grl2020,zhu2023wingnn}. As outlined in Problem~\ref{problem:1}, the key challenge of dynamic graph learning lies in designing an encoder that effectively captures intricate temporal evolution and structural changes in dynamic graphs, regardless of whether they originate from continuous-time or discrete-time data. 
% This unified approach aims to bridge the gap between different dynamic graph types, enabling a more flexible and robust model for a wide range of real-world applications.

\subsection{Discrete Fourier Transform}
Inspired by our observations in Fig.~\ref{fig:edge_features}, the frequency domain can reveal insights that may not be evident in the time domain for dynamic graph learning. Thus, we aim to analyze graph signals in the frequency domain using the Discrete Fourier Transform (DFT) for domain transformation. Specifically, given a time-domain signal represented as a sequence of temporal interactions, the DFT is defined as
\begin{equation}\label{eq:1}
    X(f) = \sum_{n=0}^{N_s-1} x(n) e^{\frac{-i2\pi fn}{N_s}},
\end{equation}
where $x(n)$ is the time-domain signal, $X(f)$~\footnote{We use the calligraphic script to denote a complex-number vector/matrix.} is a complex number, $N_s$ is the signal length, and $f$ represents the frequency components. The DFT is a bijective transformation, enabling the original sequence to be perfectly reconstructed through the Inverse DFT (IDFT):
\begin{equation}\label{eq:2}
    x(n) = \frac{1}{N_s} \sum_{f=0}^{N_s-1} X(f) e^{\frac{i 2 \pi fn}{N_s}}.
\end{equation}

The choice of DFT in UniDyG is driven by two key factors: its discrete nature, which is well-suited for digital processing, and the availability of efficient computational algorithms.

% After a snapshot is observed, it can be used to update the node representation of the snapshot
% based models for the prediction of the next snapshot (only forward pass for inference).

\begin{figure*}
\centering
\includegraphics[width=1\textwidth]{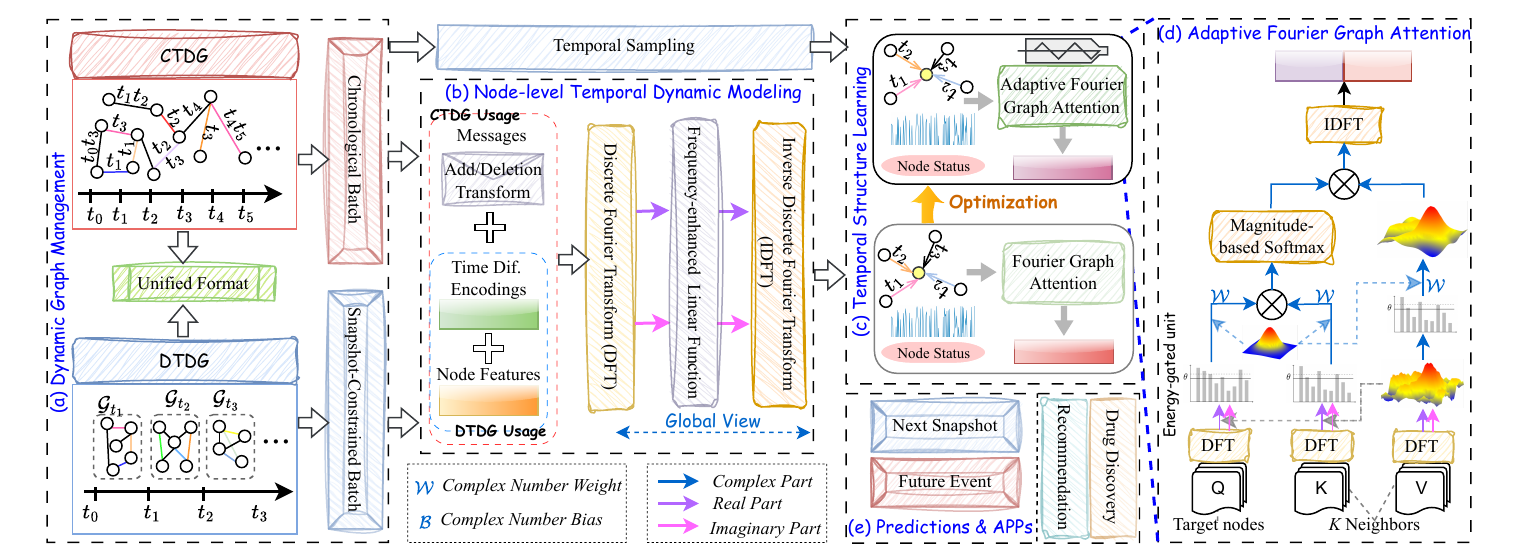}
\caption{The overview of the proposed UniDyG. Given input from CTDG or DTDG in (a), it converts the input into a unified format, together with data splits. (b) presents a temporal dynamic model, while (c) shows structure learning using our Fourier graph attention mechanisms. In (d), our adaptive FGAT is illustrated, and (e) highlights downstream tasks.}\label{fig:overview}  
\end{figure*}

\section{The Proposed Fourier Graph Attention}
\subsection{Motivation}
To achieve a unified model for both CTDGs and DTDGs, a key module is temporal structure learning, as these two types have different temporal granularities and structural evolutions.
We can observe from Figs.~\ref{fig:WIKI_F} and~\ref{fig:Alpha_F} that (1) Both CTDGs and DTDGs concentrate significant edge feature energies in low-frequency bands in the frequency domain, implying long-term smooth structure evolution underlying dynamic graphs. (2) There are few high-frequency signals including significant abrupt changes and noise, providing insights for analyzing dynamic graph signals. In contrast, (3) there are significantly different graph signals in the time domain with rapid changes in CTDGs and gradual trends in DTDGs, as evidenced in Figs.~\ref{fig:WIKI_T} and~\ref{fig:Alpha_T}. Existing structure learning approaches rely on the message propagation between neighbor nodes or process structure learning within each snapshot (based on an adjacency matrix), which fails to capture global structure evolution. Although they can capture long-term graph evolution by increasing the network layers, this also leads to computational inefficiency.

\subsection{Formulation of Fourier Graph Attention}\label{sec:4.2}
Motivated by the above observations and analysis, we explore the potential of temporal structure learning in the frequency domain and propose a novel Fourier Graph Attention (FGAT) mechanism, which can effectively model underlying structural patterns underpinning CTDGs and DTDGs. The key idea is to explore both local and global structural correlations depending on local graph structures and frequency bands. Concretely, assuming the input to the $l$-th layer consists of node $u_i$'s representation $\bm{Z}_{i}^{l-1}(t)$, the current timestamp $t$, $u_i$'s neighborhood representations $\{\bm{Z}_{i1}^{l-1}(t),\cdots, \bm{Z}_{iN}^{l-1}(t)\}$ with corresponding features (including timestamps and edge attributes). We formulate our Fourier graph attention mechanism as 
\begin{equation}
 \mathbf{Z}_{i}^{l}(t) = \mathcal{F}^{-1} \left( \sum_{j=1}^{N} \alpha_{ij} \odot \left( \mathcal{W}_v^{l} \odot \mathcal{F}\left( \bm{V}_{ij}^{l}(t) \right)\right) \right),\label{eq:3}   
\end{equation}
\begin{equation}
    \alpha_{ij} = \frac{\exp\left(\left| \mathcal{A}_{ij} \right|\right)}{\sum\limits_{m=1}^{N} \exp\left(\left| \mathcal{A}_{im}  \right| \right)},\label{eq:4}
\end{equation}
\begin{equation}
     \mathcal{A}_{ij} = \left( \mathcal{W}_q^{l} \odot \mathcal{F}\left( \bm{Q}_{i}^{l}(t) \right) \right)^{T} \odot \left( \mathcal{W}_k^{l} \odot \mathcal{F}\left( \bm{K}_{ij}^{l}(t) \right)\right).\label{eq:5}
\end{equation}

Here, $\bm{K}_{ij}^{l}(t) = \bm{V}_{ij}^{l}(t) = [\phi(t-t_j)\|\bm{e}_{ij}]$, and $\bm{Q}_{i}^{l}(t) = [\bm{Z}_{i}^{l-1}(t)\|\phi(0)]$. $\odot$ represents element-wise multiplication and $\phi(\cdot)$ is the time encoding function~\cite{DBLP:conf/iclr/CongZKYWZTM23}. $\mathcal{W}_q^{l}, \mathcal{W}_k^{l}, \mathcal{W}_v^{l}$ are learnable complex-number weight matrices for \textit{query}, \textit{key}, and \textit{value} projections. Eq.~\eqref{eq:4} denotes our magnitude-based Softmax function, where $\left| \mathcal{A}_{ij} \right| = \sqrt{ \left( \operatorname{Re}\left( \mathcal{A}_{ij} \right) \right)^2 + \left( \operatorname{Im}\left( \mathcal{A}_{ij} \right) \right)^2 }$. $\operatorname{Re}(\cdot)$ and $\operatorname{Im}(\cdot)$ denote the real and imaginary parts of complex numbers, respectively. The operation $\sum_{j=1}^{N} \alpha_{ij} \odot \left( \mathcal{W}_v^{l} \odot \mathcal{F}\left( \bm{V}_{ij}^{l}(t) \right)\right)$ in Eq.~\eqref{eq:3} is the complex-number selective aggregation for node $u_i$ from its $N$ neighbors in the frequency domain, which will be analyzed in next subsection. $\mathcal{F}(\cdot)$ and $\mathcal{F}^{-1}(\cdot)$ denote the Fourier transform and inverse Fourier transform, defined in Eqs.~\eqref{eq:1} and~\eqref{eq:2}, respectively. Last, we achieve the output representations (\textit{e.g.}, $\mathbf{Z}_{i}^{l}(t)$) from the FGAT, preserving local structure integrity and global structural evolution underlying dynamic graphs.

\subsection{Theoretical Analysis for Our FGAT}\label{sec:4.3}
Our proposed FGAT is beneficial for both continuous-time and discrete-time dynamic graph learning on four aspects: (1) \underline{Local structural correlation learning}. Our proposed FGAT employs $N$ neighborhood representations $\{\bm{Z}_{i1}^{l-1}(t),\cdots, \bm{Z}_{iN}^{l-1}(t)\}$ along with their corresponding timestamps and attributes as the input of key $\bm{K}$ and value $\bm{V}$, This enables the model to capture local correlations between a node and its neighbors, recording in coefficients, \textit{e.g.}, $\mathcal{A}_{ij}$. (2) \underline{Global structural correlation learning} lies in two-fold. First, by transforming the attribute data and time encodings into the frequency domain using the Fourier transform $\mathcal{F}(\cdot)$ in Eqs.~\eqref{eq:3} and~\eqref{eq:5}, the model can express signals as a combination of sinusoids, making it capable of detecting periodic patterns and long-term dependencies. Second, in the frequency domain, the multiplication operation $\mathcal{W}_v^{l} \odot \mathcal{F}\left( \bm{V}_{ij}^{l}(t) \right)$ in Eq.~\eqref{eq:3} can be interpreted as a form of global convolution according to~\cite{yi2024frequency}. Different $\alpha_{ij}$ can selectively emphasize certain frequency components, which is useful for handling dynamics at different time scales. The summation operation $\sum_{n=1}^{N}$ represents the aggregation of contributions from $N$ neighboring nodes, obtaining a high-quality frequency-domain representation. Therefore, based on the complex-number selective aggregation in Eq.~\eqref{eq:3}, our FGAT allows each neighbor to influence the entire spectrum, thus capturing global structural correlations. (3) \underline{Temporal coherence}. Our FGAT maintains temporal coherence by demonstrating Lipschitz continuity \textit{w.r.t.} time translations. This property ensures that small temporal shifts of the input in the time domain lead to proportionally small changes in the output representations, resulting in consistent temporal embeddings. Consequently, FGAT can generate temporally coherent representations of dynamic graphs, which will be supported by the Theorem~\ref{theorem:1}.
(4) \underline{Universal approximation}. Our FGAT can learn to represent arbitrarily complex transformations of graph signals in the frequency domain, making it versatile for different types of graph structures and signal properties across both CTDGs and DTDGs. To support this point, we present the Theorem~\ref{theo:1}.

\tcbset{colback=lightgray!15!white, colframe=white, left=1mm, right=1mm, top=1mm, bottom=1mm}
\begin{tcolorbox}
    \begin{theorem}[Temporal Coherence]\label{theorem:1}
        The Fourier Graph Attention (FGAT) mechanism is Lipschitz continuous with respect to time translations in the input signals. That is, there exists a constant \( L > 0 \) such that for any two timestamps \( t \) and \( t' \), the output representations satisfy:
\[
\left\| \mathbf{Z}_{i}^{l}(t) - \mathbf{Z}_{i}^{l}(t') \right\| \leq L | t - t' |,
\]
where $\left\|\cdot\right\|$ is the norm and $|\cdot|$ denotes the absolute value.
    \end{theorem}
\end{tcolorbox}
\begin{proof}
To prove the Lipschitz continuity of our FGAT, we need to demonstrate that each operator of the FGAT mechanism is continuous concerning the timestamp \( t \). First, for any timestamps $t, t'$, the input of the query $\bm{Q}_{i}^{l}(t)$, key $\bm{K}_{ij}^{l}(t)$, and value $\bm{V}_{ij}^{l}(t)$ are composed of node features and time encodings. We take the query as an example, involving the time encoding function $\phi(\cdot)$, which can preserve the Lipschitz continuity~\cite{brigham1988fast}. Thus we achieve:
\begin{equation}\label{eq:1.6}
    \| \bm{Q}_{i}^{l}(t) - \bm{Q}_{i}^{l}(t') \| = \| \bm{Z}_{i}^{l-1}(t) - \bm{Z}_{i}^{l-1}(t') \| \leq L_z | t - t' |,
\end{equation}
where $\phi(0)$ is the constant. We leverage Fourier transform $\mathcal{F}(\cdot)$ for the input $\bm{Q}_{i}^{l}(t)$, $\bm{K}_{ij}^{l}(t)$, and $\bm{V}_{ij}^{l}(t)$, which is a linear operator, thereby preserving the Lipschitz continuity. Then we compute the attention weights $\alpha_{ij}$ by Eqs.~\eqref{eq:4} and~\eqref{eq:5}. Since the element-wise multiplication preserves Lipschitz continuity, we achieve 
\begin{equation}\label{eq:1.7}
    \| \mathcal{A}_{ij}(t) - \mathcal{A}_{ij}(t') \| \leq L_A | t - t' |,
\end{equation}
The magnitude function \(| \cdot |\) is Lipschitz continuous with constant $1$. The exponential function is smooth and Lipschitz continuous on bounded domains. Since \(\mathcal{A}_{ij}(t)\) are bounded (due to bounded inputs and weights), there exists a constant \(L_\alpha\) such that:
 \begin{equation}\label{eq:1.8}
      \| \alpha_{ij}(t) - \alpha_{ij}(t') \| \leq L_{\alpha} | t - t' |.
 \end{equation}
Then, we compute the output representations based on Eq.~\eqref{eq:3}, where each term in the sum is Lipschitz continuous. Together with the Eqs.~\eqref{eq:1.6}-~\eqref{eq:1.8}, we can achieve 
\begin{equation}
    \left\| \mathbf{Z}_{i}^{l}(t) - \mathbf{Z}_{i}^{l}(t') \right\| \leq L | t - t' |,
\end{equation}
where \(L\) is a constant depending on constants \(L_z\), $L_A$, $L_{\alpha}$, the norms of the weights \(\mathcal{W}_q^{l}\), \(\mathcal{W}_k^{l}\), \(\mathcal{W}_v^{l}\), and the number of neighbors \(N\). The proof is completed.
\end{proof}

% Equivariance to time translation means FGAT consistently models the temporal dynamics of graphs

\tcbset{colback=lightgray!15!white, colframe=white, left=1mm, right=1mm, top=1mm, bottom=1mm}
\begin{tcolorbox}
\begin{theorem}[Universal Approximation]\label{theo:1}
    Give a dynamic graph $G$ with $|U|$ nodes and attribute matrix $X$. Consider the space of continuous functions $\mathcal{F} = \{f: \mathbb{R}^D\rightarrow\mathbb{R}^d\}$ operating on the frequency domain of graph signals. Our proposed Fourier Graph attention mechanism possesses the universal approximation property in the frequency domain. That is, for any function $f\in \mathcal{F}$ and any $\epsilon > 0$, there exists a Fourier graph attention network $\operatorname{FGAT}(\cdot)$ with $L$ layers such that:
    \begin{equation}
    \sup_{\bm{x} \in \bm{X}} \left\| \text{FGAT}(\bm{x}) - f(\mathcal{F}(\bm{x})) \right\| < \epsilon,
\end{equation}
where $\mathcal{F}(x)$ denotes the Fourier transform of the graph signal $x$, and $\left\|\cdot\right\|$ is the norm.
\end{theorem}
\end{tcolorbox}

\begin{proof}
Let $f:\mathbb{R}^D \to \mathbb{R}^d $ be a continuous function in the frequency domain of graph signals and $\epsilon > 0$ be an arbitrary positive real number.
Since we are dealing with graph signals, we can assume that the frequency domain representation is bounded. Let $\Omega \subset \mathbb{R}^d$ be a compact subset that contains all possible Fourier transforms of the graph signals.
Based on the Stone-Weierstrass theorem~\cite{stone1948generalized}, there exists a finite set of simple functions $\{g_i : \Omega \to \mathbb{R}\}_{i=1}^m$ such that:
\begin{equation}
\sup_{\bm{x} \in \Omega} \left\|f(\bm{x}) - \sum_{i=1}^m c_i g_i(\bm{x})\right\| < \frac{\epsilon}{2}.
\end{equation}
where $c_i$ are appropriate coefficients.

Next, we construct a Fourier graph attention network that approximates these simple functions. Consider a single layer of the network:
\begin{equation}
\mathbf{Z}_{i}(t) = \mathcal{F}^{-1} \left( \sum_{j=1}^{N} \alpha_{ij} \odot \left( \mathcal{W}_v \odot \mathcal{F}\left( \bm{V}_{ij}(t) \right)\right) \right),
\end{equation}
where $\alpha_{ij}$ is computed by the Eq.~\eqref{eq:4}.
We can show that for each simple function $g_i$, there exists a set of weights $\mathcal{W}_q$, $\mathcal{W}_k$, and $\mathcal{W}_v$ such that:
\begin{equation}
\left\|g_i(\mathcal{F}(\bm{x})) - \bm{Z}_t(t)\right\| < \frac{\epsilon}{2m}
\end{equation}
This is achieved by choosing appropriate $\mathcal{W}_q$ and $\mathcal{W}_k$ to compute attention coefficients $\alpha$ that focus on relevant frequency components, and $\mathcal{W}_v$ to mix these components.

By combining multiple of these layers (each approximating one $g_i$), we can construct a network that approximates the sum $\sum_{i=1}^m c_i g_i(\bm{x})$. Based on the above analysis and given the input $\bm{x}' = \mathcal{F}(\bm{x})$, we have
\begin{equation}
    \begin{split}
    &\left\| f (\bm{x}') - \text{FGAT}(\bm{x})\right\|\leq \\
&\|f(\bm{x}') - \sum_{i=1}^m c_i g_i(\bm{x}')\| + \|\sum_{i=1}^m c_i g_i(\bm{x}') - \text{FGAT}(\bm{x})\| \\
&<\frac{\epsilon}{2} + \frac{\epsilon}{2} = \epsilon.
    \end{split}
\end{equation}
The final operator, inverse Fourier transform $\mathcal{F}^{-1}(\cdot)$, ensures that the approximation in the frequency domain translates to the time domain, preserving the approximation property.

This construction shows that for any continuous function $f$ on the frequency domain of graph signals and any $\epsilon > 0 $, we can construct a Fourier Graph Attention network that approximates $f$ within $\epsilon$ accuracy.
\end{proof}

\subsection{Further Optimization}
We observe in Fig.~\ref{fig:edge_features} that temporal noise is ubiquitous in real-world dynamic graphs, manifesting as random or irrelevant short-term fluctuations in graph structure or node/edge attributes. Existing dynamic graph methods often overlook the temporal noise issue, while time-domain models typically lack mechanisms to distinguish between meaningful abrupt changes and noise, resulting in sub-optimal performance and reduced robustness. To address this, we first analyze temporal noise in the frequency domain, where it corresponds to high-frequency components~\cite{DBLP:conf/icml/Eldele0C0024}—some of which may still carry valuable information. We then optimize our proposed Fourier graph attention by designing an energy-gated unit $\mathcal{L}_{G}(\cdot)$, which selectively retains high-frequency components based on their energy. Concretely, given the dynamic graph features $\bm{X}_{c}$ and their frequency transform $\bm{P} = \mathcal{F}(\bm{X}_{c})$ as input, we formulate the energy-gated unit $\mathcal{L}_{G}(\cdot)$ as
\begin{equation}
\mathcal{L}_{G}(\bm{P}) = \mathcal{W}_{G}\odot(\bm{P} \odot \bm{M}) + \mathcal{B}, \;\bm{M} = \mathbb{I}(\|\bm{P}\|^2 < \theta),
\end{equation}
where $\mathbb{I}(\cdot)$ is the indicator function, returning $1$ if true, $0$ otherwise. $\mathcal{W}_{G}$ and $\mathcal{B}$ are learnable complex-number parameters. $\theta$ is a hyperparameter threshold for energy gating. By selectively retaining high-frequency components based on their energy, the mechanism can preserve crucial high-frequency signals that may carry important information about abrupt changes in the graph structure. By explicitly addressing the noise issue, the energy-gated unit contributes to overall model robustness, helping to prevent overfitting to short-term fluctuations or measurement errors. 

Then, we integrate it into our proposed Fourier graph attention by replacing the linear projections with our energy-gated unit, thereby constructing an adaptive Fourier graph attention mechanism, referred to as FGAT\_N. We formulate it as 
\begin{align}
&\mathbf{Z}_{i}^{l}(t) = \mathcal{F}^{-1} \left( \sum_{j=1}^{N} \alpha_{ij} \odot \mathcal{L}_{G}\left(\mathcal{F}\left( \bm{V}_{ij}^{l}(t) \right)\right) \right),\label{eq:12}\\
\mathcal{A}_{ij} &= \mathcal{L}_{G}\left( \mathcal{F}\left( \bm{Q}_{i}^{l}(t) \right) \right)^{T} \odot \mathcal{L}_{G}\left( \mathcal{F}\left( \bm{K}_{ij}^{l}(t) \right)\right),\label{eq:13}
\end{align}
where $\alpha_{ij}$ is computed in the same manner as in Eq.~\eqref{eq:4}.
According to Eqs.~\eqref{eq:12} and~\eqref{eq:13}, the optimized FGAT\_N can adaptively filter out high-frequency noise from the attributes and structures based on the energy, thereby improving model robustness. Besides, the four advantages of our Fourier graph attention mechanism mentioned in Section~\ref{sec:4.3} remain intact. Fig.~\ref{fig:overview}d illustrates our adaptive Fourier graph attention mechanism.

% further optimize our proposed Fourier graph attention by designing an energy-gated unit. In the frequency domain, temporal noise corresponds to high-frequency components, although some of these components may contain valuable information. Here, we design a novel energy-gated unit $\mathcal{L}_{EG}(\cdot)$ that selectively retains high-frequency components based on their energy. 

\section{The Proposed UniDyG}
\subsection{Overview}
Building on our frequency domain observations and our proposed Fourier graph attention mechanisms, we present UniDyG, a unified framework in the frequency domain for modeling both CTDGs and DTDGs, as shown in Fig.~\ref{fig:overview}. UniDyG excels in capturing long-term dependencies and optimizing model robustness. It comprises three components: a node-level temporal dynamic model, a Fourier structure encoder, and a unified training pipeline. Concretely, to model the continuous and discrete temporal dynamics, we leverage the frequency-enhanced linear function to model temporal dependencies, together with different message extractors, which can provide a global view and efficiently model long-term temporal dependencies. For the structure modeling, we aim to capture both the smooth evolution characteristic of CTDGs and the abrupt shifts typical in DTDGs, in which we propose a dual Fourier structure encoder for time-oriented structure learning and attribute-oriented structure learning, respectively. We develop a streamlined yet effective training pipeline to minimize information loss and prevent information leakage in dynamic graph learning. 

% We provide theoretical proof that UniDyG can approximate any continuous function on graph signals in the frequency domain. This makes it adaptable to various graph structures and signal properties, encompassing both CTDGs and DTDGs. 

\subsection{Temporal Dynamic Modeling}
To capture node dynamic evolution, dynamic graph methods typically generate the state vector $\bm{S}(t)$ for each node at time $t$ and update node states based on evolving temporal interactions, paving the way for subsequent dynamic node representation. Existing methods typically employ complex sequence models~\cite{DBLP:conf/kdd/KumarZL19,DBLP:conf/wsdm/SankarWGZY20,tgn_icml_grl2020,DBLP:journals/pacmmod/LiSCY23} or simple linear models~\cite{DBLP:conf/iclr/CongZKYWZTM23,TimeSGN} to update the node state, both of which struggles with long-term temporal dependencies in the time domain, failing to model the dynamics for discrete-time dynamic graphs. 
In this paper, we employ the frequency-enhanced linear function that is designed for the complex number of frequency components, to effectively capture temporal dependencies with the global view. Concretely, we first record node messages including edge/node addition/deletion or edge feature transformations by the message function~\cite{tgn_icml_grl2020} and then obtain a message vector for each node, \textit{i.e.}, $\bm{m}(t)$, exclusively for CTDGs. In case of a node-wise event $u_i(t)$, a single message can be computed for the node involved in the event: $\bm{m}_i(t) = \operatorname{msg}_{n}(\bm{S}_i(t^{-}), t, u_i(t))$, where $\operatorname{msg}_{n}(\cdot)$ is the node-level message function. Given the historical node state $\bm{S}_i(t^{-})$, current messages $\bm{m}_i(t)$, node features $\bm{x}_i(t)$ and timestamps as the input, we formulate our update function by
% \begin{equation}\label{eq:11}
% \bm{s}_i(t) = \operatorname{FourierMLP}(\bm{s}_i(t^{-})\parallel\operatorname{LN}[\bm{m}_i(t)\parallel \phi(t-t^-)])）,
% \end{equation}
\begin{equation}
  \mathcal{S}'_i(t) = \mathcal{W}_d\odot \left (\mathcal{F}\left ([\bm{S}_i(t^{-})\|\bm{m}_i(t)\| \phi(t-t^-)\|\bm{x}_i(t)])\right )\right ),  
\end{equation}
\begin{equation}
    \bm{S}_i(t) = \sigma\left (\mathcal{F}^{-1}\left (\mathcal{S}'_i(t)\right ) \right ),
\end{equation}
where $\mathcal{W}_d$ is the complex-number weights. The proposed temporal dynamic model, with its combination of time-domain and frequency-domain processing, is well-suited for capturing long-term temporal dependencies in both discrete and continuous time scenarios. This is because frequency domain analysis can reveal periodic patterns over long time scales. For the DTDGs, we don't use the message function for batch processing. This is because each snapshot would be split into multiple batches. Such a message function may interfere with the setting of discrete-time dynamic graphs, incurring extra dynamics within each snapshot. Our approach provides a flexible framework that can adapt to irregular event timings and identify long-term temporal dependencies in the frequency domain across CTDGs and DTDGs.

% The frequency domain processing offers a powerful tool for identifying long-term patterns in both cases, potentially overcoming some limitations of purely time-domain approaches in capturing distant temporal relationships.
% The effectiveness in capturing long-term dependencies ultimately depends on the specific nature of the data and the length of the sequences being analyzed. 

\subsection{FGAT and FGAT\_N for Temporal Structure Learning}
Building on our proposed Fourier graph attention, we leverage it to model complex temporal structures for generating node embeddings. For dynamic graphs, each edge has multiple types of features, including timestamps and attribution information. So, we apply our proposed FGAT and FGAT\_N into the divided temporal message passing paradigm~\cite{TimeSGN} for complex structure learning, which enables expressing individual characteristics for temporal embeddings without any masking. Considering the noise issue, we leverage our FGAT for time-oriented structure learning while employing our FGAT\_N for attribute-oriented structure learning, which can preserve complete time granularity and filter out noise in attribute information. Later, these two embeddings are effectively integrated to generate aggregated neighborhood embeddings, together with previous temporal embeddings for final temporal embeddings. We formulate them as
\begin{align}
&\bm{Z}_{i}^{l}(t) = \bm{Z}_{i}^{l-1}(t)+\operatorname{ReLU}(\operatorname{FFN}(\bm{Z}_{i,\tau}^{l}(t)\| \bm{Z}_{i,\epsilon}^{l}(t))),\label{eq:16}\\
&\bm{Z}_{i,\tau}^{l}(t)= \operatorname{FGAT}(\bm{Q}_{i,\tau}^{l}(t), \bm{K}_{i,\tau}^{l}(t), \bm{V}_{i}^{l}(t)), \label{eq:17}\\
&\bm{Z}_{i,\epsilon}^{l}(t) = \operatorname{FGAT\_N}(\bm{Q}_{i,\epsilon}^{l}(t), \bm{K}_{i,\epsilon}^{l}(t), \bm{V}_{i,\epsilon}^{l}(t)),\label{eq:18}
\end{align} 
where $\operatorname{ReLU}(\cdot)$ is the activation function. $\bm{K}_{i,\tau}^{l}(t) = \bm{V}_{i,\tau}^{l}(t) = [\phi(t-t_1)\| \cdots \|\phi(t-t_N)]$, $\bm{Q}_{i,\tau}^{l}(t) = [\phi(0)]$, $\bm{K}_{i,\epsilon}^{l}(t) = \bm{V}_{i,\epsilon}^{l}(t) = [\bm{e}_{i1}(t_1)\|\bm{S}_1(t) \cdots \|\bm{e}_{iK}(t_N)\|\bm{S}_N(t)]$, and $\bm{Q}_{i,\epsilon}^{l}(t) = [Z_{i}^{l-1}(t')\|\bm{S}_i(t)]$. We accomplish structure learning with timestamps based on Eq.~\eqref{eq:17}, capturing local and global correlations to generate time-based embeddings $\bm{Z}_{i,\tau}^{l}(t)$. Based on Eq.~\eqref{eq:18}, we filter out the temporal noise of attribute information and accomplish structure learning with edge attributes and node status, generating attribute-based embeddings $\bm{Z}_{i,\epsilon}^{l}(t)$. Both embeddings are integrated to generate high-quality temporal embeddings $\bm{Z}_{i}^{l}(t)$ based on Eq.~\eqref{eq:16}.

\subsection{Unified Training Pipeline}\label{sec:4.4}
We present a unified training pipeline that accommodates both CTDGs and DTDGs considering their characteristics, as shown in Fig.~\ref{fig:overview}. This pipeline includes the neighbor sampling and the training procedure.

\subsubsection{Neighbor Sampling}
As discussed in Section~\ref{sec:4.2}, selecting recent neighbors is essential for the effectiveness of our Fourier graph attention. However, strategies for neighbor selection differ significantly between CTDGs and DTDGs. CTDG-specific methods typically focus on temporal sampling for time-aware neighbor selection, while DTDG-specific methods rely on constructing adjacency matrices within each snapshot. In our streaming setting, where data is continuously arriving in real time, constructing full adjacency matrices becomes impractical and can easily lead to information leakage. This occurs when future data is unintentionally used to learn about the current state of the node, resulting in inaccurate results. To address this, we employ temporal sampling~\cite{zhou2022tgl} for neighbor selection in DTDGs by limiting the time window, which avoids the need for full adjacency matrix construction. Concretely, we sample a fixed number of neighbors (\textit{e.g.}, $N$) based on their temporal proximity for subsequent temporal structure learning, which is effective for both CTDGs and DTDGs.

\subsubsection{Training Procedure}\label{sec:5.4.2}
For our training procedure, we adopt mini-batch processing, as it is commonly used in CTDG learning. However, directly applying this approach in a streaming setting for DTDGs can result in information leakage. That is, the model may use data from snapshots up to $G(t)$ to predict edges for the same time $t$. Furthermore, existing DTDG-specific methods typically treat each snapshot independently during model training. To ensure compatibility with mini-batch processing, we propose a straightforward yet efficient training strategy for DTDGs, consisting of two key components.
\begin{itemize}
\item \textit{Training/Validation/Test Split.}  We partition the DTDG into three sets to ensure temporal consistency and prevent future information leakage. Training Set: Snapshots from $G(1)$ to $G(T_1)$, Validation Set: Snapshots from $G(T_1+1)$ to $G(T_2)$, and Test Set: Snapshots from $G(T_2+1)$ to $G(T)$, where $1 < T_1 < T_2 < T$, and $T$ is the total number of snapshots. Crucially, we enforce that the model only utilizes information from past snapshots when processing any given snapshot, thereby preventing any leakage of future information.

\item \textit{Mini-Batch Construction.} For each snapshot $G(t)$, we employ an edge-sampling approach to construct mini-batches: We define a fixed batch size $B$ (\textit{e.g.}, $B = 600$) for consistency across training. Instead of sampling nodes, we sample edges from the snapshot to capture active interactions within the graph. For a snapshot with $|E(t)|$ edges, we create $\lfloor |E(t)| / B \rfloor$ full mini-batches of size $B$. The remaining $|E(t)| \bmod B$ edges form a final, potentially smaller, mini-batch. We do not pad the final batch, thereby preserving the true graph structure and avoiding the introduction of artificial data.
\end{itemize}

This approach ensures that each edge in a snapshot is processed exactly once per epoch, maintaining the integrity of the graph structure while allowing for efficient batch processing. By combining training-test splitting with mini-batch construction, our UniDyG provides a robust framework for both CTDG and DTDG learning that respects temporal causality and captures the dynamic nature of graph interactions.

\subsection{Complexity Analysis}
The computational cost of our proposed UniDyG is mainly attributed to two modules: the temporal dynamic model and the temporal structural encoder. For temporal dynamic modeling, the complexity is \(O(D_d \cdot \log(D_d))\), where \(D_d\) is the combined dimension of its inputs \(\bm{S}_i(t^{-}), \bm{m}_i(t), \phi(t - t^-)\), and \(\bm{x}_i(t)\). In the temporal structural encoder, our Fourier graph attention mechanism has a complexity of \(O(L \cdot N \cdot d \cdot \log(d))\), where \(d\) is the dimension of \(\bm{Z}_{i}^{l}(t)\), \(N\) is the number of neighbors, and \(L\) represents the network layers (set to $1$ in this paper). Thus, the overall complexity of UniDyG is \(O(N \cdot d \cdot \log(d) + D_d \cdot \log(D_d))\).

In summary, our UniDyG offers several advantages: (1) UniDyG is designed to accurately model temporal structures for both CTDGs and DTDGs, tracking smooth evolutions in continuous time and detecting abrupt changes in discrete time. This is because our Fourier structure encoder can capture both local and global structural correlations based on different frequency bands and recent neighbors. (2) UniDyG is highly robust to noise and can maintain stable performance even in noisy scenarios since we integrate our energy-gated unit into our Fourier graph attention for temporal structure learning. (3) Our UniDyG can scale to large dynamic graphs by avoiding complex sequence models and multi-layer network stacking as well as relying instead on a simple and effective training pipeline. This makes it suitable for various real-world applications.

\section{Experiments}
\begin{table}[t]
  \caption{Dataset statistics. ``Unix": Unix timestamp.}
  \label{tab:1}
  % \tiny
  \resizebox{\linewidth}{!}{
  \begin{tabular}{|c|c|c|ccccc|}
    \Xhline{1.5pt}
    \rowcolor{gray!40}Type &Alias& Dataset & $|U|$ & $|E|$ & $\max(t)$ &Granularity& $d_e$  \\
    \hline
    \hline 
    \multirow{4}{*}{\rotatebox{90}{\centering DTDG}} 
    &BA& BitcoinAlpha &3,783 &24,186 &226 &Weekly & 1\\
    &BO& BitcoinOTC & 5,881 &35,592 & 262& Weekly & 1\\
    &UCI& UCI & 1,899 &59,835 &28 & Weekly& 1\\ 
    &SO& StackOverFlow & 2,601,977 &63,497,050 &92 & Monthly & - \\
    \hline
    \hline
    \multirow{5}{*}{\rotatebox{90}{\centering CTDG}} 
    &WK& Wikipedia & 9,227 &157,474 &2.7e6 &Unix& 172\\
    % &MOOC & 7,144 & 411,749 &2.6e6 & Unix&4\\
    &AU& AskUbuntu &159,316  &964,437 &2.3e8 &Unix &- \\
    &LF& LastFM &1,979  & 1,293,103& 1.3e8& Unix &- \\
    &SU& SuperUser & 194,085 &1,443,339 &2.4e8 &Unix &-\\
    &WT& WIKITALK & 1,140,149 &7,833,140 &2.0e8 &Unix &-\\
    \hline
  \end{tabular}}
\end{table}

\subsection{Experimental Setting}
\subsubsection{Datasets} Table~\ref{tab:1} summarizes the datasets used in experiments, where nine datasets are benchmarks for evaluating dynamic graph learning. Specifically, four datasets are the discrete-time dynamic graphs collected from~\cite{DBLP:conf/kdd/YouDL22} and five continuous-time dynamic graphs are collected from~\cite{zhou2022tgl,paranjape2017motifs}. We chronologically split
all the graphs in Table~\ref{tab:1} into training ($70\%$), validation ($15\%$), and test ($15\%$) sets, where we split the DTDG according to the approach in Section~\ref{sec:5.4.2} to avoid the information leakage. To test model
performance on unseen nodes, we randomly sample $10\%$ of nodes from each dataset and mask them during model training as in~\cite{tgn_icml_grl2020,DBLP:journals/pacmmod/LiSCY23}. 
% For the DTDG learning, we make the prediction over future snapshots.

\begin{table*}[t]
	\caption{Overall performance comparison on five CTDG datasets (\% is omitted) in transductive and inductive settings. The best results and second best are highlighted in blue and light blue, respectively.}
	\label{tab:2}
	\centering
 \renewcommand{\arraystretch}{1.2}
 \resizebox{\linewidth}{!}{
	\begin{tabular}{|c|c|c|cccccccc|cc|}
     \Xhline{1.5pt}
		\rowcolor{gray!40}Setting& Metric& Dataset & TGN &Zebra &Orca &GraphMixer &DyGFormer & TimeSGN & FreeDyG &CNEN &DGNN-GRU &UniDyG \\
  \hline
  \hline 
\multirow{10}{*}{\rotatebox{90}{\centering Transductive}} &\multirow{5}{*}{\centering AUC ($\uparrow$)} 
  &WK&98.23$\pm$0.2 &97.14$\pm$0.1&98.60$\pm$0.1&96.92$\pm$0.0 &98.91$\pm$0.0 &\underline{99.59$\pm$0.0} &99.41$\pm$0.0 &99.01$\pm$0.0 &90.76$\pm$0.8  &\textbf{99.71$\pm$0.0} \\
  &&AU&45.58$\pm$0.8&58.11$\pm$0.4 &60.03$\pm$0.6&81.12$\pm$0.5 &90.31$\pm$0.2 &\underline{94.23$\pm$0.1} &70.74$\pm$0.8&69.82$\pm$0.2&70.21$\pm$0.5  &\textbf{98.16$\pm$0.1} \\
  &&LF&78.47$\pm$0.1&79.93$\pm$0.1&81.37$\pm$0.6&73.53$\pm$0.2 &93.05$\pm$0.2  &\underline{95.51$\pm$0.0} &89.71$\pm$0.2 &87.03$\pm$0.1 &84.50$\pm$0.3  &\textbf{98.43$\pm$0.0} \\
  &&SU&49.82$\pm$0.5&49.15$\pm$0.4&60.02$\pm$0.9&86.55$\pm$0.7 &87.56$\pm$0.6 &\underline{93.31$\pm$0.2} &69.67$\pm$1.2&73.07$\pm$0.1 &48.83$\pm$1.3  &\textbf{97.85$\pm$0.2} \\
  &&WT&-&-&-&86.87$\pm$0.4 &89.04$\pm$0.6 &\underline{95.39$\pm$0.0} &-&36.17$\pm$0.5&-  &\textbf{97.04$\pm$0.0} \\
   \cline{2-13} 
   &\multirow{5}{*}{\centering AP ($\uparrow$)} 
  &WK&98.14$\pm$0.1 &98.22$\pm$0.1 &98.59$\pm$0.1 &97.17$\pm$0.1 &98.82$\pm$0.0 &\textbf{99.86$\pm$0.0} &99.26$\pm$0.0&99.09$\pm$0.0 &90.16$\pm$0.6 &\underline{99.72$\pm$0.0}  \\
&&AU&48.11$\pm$1.3 &56.21$\pm$0.6&53.83$\pm$0.5 &78.06$\pm$0.1 &91.78$\pm$0.6 &\underline{94.51$\pm$0.4} &67.55$\pm$0.8 &79.54$\pm$0.3 &68.34$\pm$0.6 &\textbf{98.28$\pm$0.1} \\
  &&LF&85.76$\pm$0.3 &78.28$\pm$0.4 &78.69$\pm$0.4 &75.64$\pm$0.6 &92.07$\pm$0.3 &\underline{96.21$\pm$0.0} &90.92$\pm$0.2 &88.91$\pm$0.1 &83.36$\pm$0.4 &\textbf{98.18$\pm$0.0} \\
  &&SU&93.18$\pm$0.4&49.57$\pm$0.5&67.42$\pm$0.7 &86.26$\pm$0.8 &89.07$\pm$0.6 &\underline{93.40$\pm$0.3} &66.80$\pm$1.0&82.39$\pm$0.0 &49.58$\pm$1.1 &\textbf{98.02$\pm$0.1} \\
  &&WT&-&-&- &81.54$\pm$0.5 &85.01$\pm$1.0 &\underline{95.68$\pm$0.1} &-&57.71$\pm$0.7&-&\textbf{97.39$\pm$0.0} \\
   \hline 
   \hline
   \multirow{10}{*}{\rotatebox{90}{\centering Inductive}} &\multirow{5}{*}{\centering AUC ($\uparrow$)} 
  &WK&97.72$\pm$0.1&98.72$\pm$0.1&98.14$\pm$0.1&96.30$\pm$0.3& 98.48$\pm$0.2 &\underline{99.14$\pm$0.0} &99.01$\pm$0.0 &98.23$\pm$0.1 &84.08$\pm$1.1 &\textbf{99.29$\pm$0.0} \\
  &&AU&50.00$\pm$0.6&57.93$\pm$0.3&66.60$\pm$0.4 &73.75$\pm$0.9 &\underline{80.57$\pm$0.4 }&76.08$\pm$0.6 &70.71$\pm$0.5&66.93$\pm$0.4 &56.72$\pm$0.9 &\textbf{87.09$\pm$0.1}\\
  &&LF&82.61$\pm$0.2&95.30$\pm$0.1&90.88$\pm$0.3 &80.37$\pm$0.3 &94.08$\pm$0.1 &\underline{97.72$\pm$0.0} &92.39$\pm$0.2 &91.62$\pm$0.0 &80.90$\pm$0.3 &\textbf{99.89$\pm$0.0}\\
  &&SU&50.03$\pm$0.6&44.68$\pm$0.5&67.19$\pm$0.5 &76.10$\pm$0.1 &\underline{77.56$\pm$1.0} &76.26$\pm$0.7 &66.16$\pm$0.9&68.89$\pm$0.3 &66.82$\pm$0.7&\textbf{89.16$\pm$0.1}\\
  &&WT&-&-&-&84.71$\pm$0.4 &\underline{85.94$\pm$0.6} &74.53$\pm$0.3&-&36.03$\pm$0.6&-&\textbf{86.75$\pm$0.2}\\
   \arrayrulecolor{black}\cline{2-13}  
   &\multirow{5}{*}{\centering AP ($\uparrow$)} 
  &WK&97.57$\pm$0.1 &98.02$\pm$0.2&97.20$\pm$0.2 &96.65$\pm$0.2&98.59$\pm$0.1 &\underline{99.26$\pm$0.0} &98.59$\pm$0.1 &98.37$\pm$0.0 &88.54$\pm$0.8 &\textbf{99.41$\pm$0.0} \\
  &&AU&52.47$\pm$0.8 &56.35$\pm$0.5&60.77$\pm$0.1  &69.25$\pm$0.6 &\underline{84.82$\pm$0.7} &82.17$\pm$0.7 &67.39$\pm$1.1&76.32$\pm$0.0 &60.33$\pm$1.6 &\textbf{89.34$\pm$0.1}\\
  &&LF&77.20$\pm$0.3 &94.60$\pm$0.3 & 90.75$\pm$0.3 &82.11$\pm$0.3 &94.23$\pm$0.5 &\underline{98.39$\pm$0.0} &93.30$\pm$0.5 &93.01$\pm$0.1&79.11$\pm$0.3&\textbf{99.76$\pm$0.0}\\
  &&SU&81.76$\pm$0.6&49.16$\pm$0.5 &70.27$\pm$0.7  &76.60$\pm$1.0 &\underline{82.67$\pm$0.9} &82.17$\pm$0.3&64.29$\pm$1.0&78.51$\pm$0.1 &62.40$\pm$0.9 &\textbf{90.75$\pm$0.3}\\
  &&WT&-&-&-&78.52$\pm$0.6 &\underline{80.32$\pm$0.8} &79.03$\pm$0.9&-&56.37$\pm$0.8&-&  \textbf{87.72$\pm$0.4}\\
   \hline 
\end{tabular}}
\end{table*}
% GraphMixer DyGFormer FreeDyG

\subsubsection{Compared Methods} We compare our proposed UniDyG with sixteen SOTA dynamic graph representation approaches, including eight CTDG-specific models, five DTDG-specific models, and three unified models.
\begin{itemize}
    \item \textit{Unified Models.} DGNN~\cite{DBLP:journals/pvldb/ZhengWL23} decouples the temporal propagation and prediction processes on dynamic graphs. Concretely, it designs the heuristic dynamic propagation algorithm for both CTDGs and DTDGs and leverages three sequence models (\textit{i.e.}, GRU, LSTM, and Transformer) for the prediction tasks, which are denoted as DGNN-GRU, DGNN-LSTM, and DGNN-TF.
    \item \textit{CTDG-Specific Models.} TGN~\cite{tgn_icml_grl2020} provides a generic framework including the memory and embedding modules, which unifies prior works as special cases. Zebra~\cite{DBLP:journals/pvldb/LiSCY23}, Orca~\cite{DBLP:journals/pacmmod/LiSCY23} and TimeSGN~\cite{TimeSGN} leverage different accelerated techniques, enhancing model efficiency and scalability. GraphMixer~\cite{DBLP:conf/iclr/CongZKYWZTM23} proposes a simple dynamic graph learning method based on the MLP-Mixer. DyGFormer~\cite{yu2023towards} and CNEN~\cite{cheng2024co} explore the co-neighbor encoding techniques to improve embedding quality. FreeDyG~\cite{tian2023freedyg} firstly designs adaptive filtering in the frequency domain to filter out irrelevant information, where structure learning is modeled in the time domain.
    \item \textit{DTDG-Specific Models.} EvolveGCN~\cite{pareja2020evolvegcn} employs an RNN to update the internal GNN parameters between snapshots, using two different temporal encoders, LSTM and GRU, which are referred to as EvolveGCN-O and EvolveGCN-H, respectively. Roland~\cite{DBLP:conf/kdd/YouDL22} and WinGNN~\cite{zhu2023wingnn} leverage meta-learning to capture temporal dependencies, together with static GNNs for structure learning. HGWaveNet~\cite{DBLP:conf/www/BaiNZZY23} leverages hyperbolic space for complex structure learning, which excels at modeling hierarchical structures.
\end{itemize}

\subsubsection{Evaluation Metrics} We evaluate the quality of temporal embeddings on dynamic link prediction tasks for both CTDGs and DTDGs across various experimental settings. Specifically, for CTDGs, we measure link prediction performance in two scenarios: the transductive and inductive settings. In line with prior work, we regard the task as a binary classification problem. For the training/test set, we randomly sample an equal number of non-existing edges to serve as negative edges. We use Average Precision (AP) and Area Under the Curve (AUC)~\cite{tgn_icml_grl2020,DBLP:journals/pvldb/LiSCY23,DBLP:journals/pacmmod/LiSCY23} as evaluation metrics. For DTDGs, we evaluate link prediction performance over future snapshots under two settings: classification and ranking. In the classification setting, we sample one negative destination node for each temporal link node pair and use AUC as the evaluation metric. In the ranking setting, we sample $100$ negative destination nodes for each source node and rank the positive temporal link node pairs higher than the $100$ negative destination nodes, using Mean Reciprocal Rank (MRR) as the metric~\cite{DBLP:conf/kdd/YouDL22,zhu2023wingnn}.
Regarding efficiency, we measure the overall training time of dynamic graph representation methods, as selected baselines employ different training pipelines for model training. We repeat the experiment with $5$ random seeds and report mean results and standard deviation. If a method fails to return results within three days or runs Out Of Memory (OOM) on a given dataset, we exclude it from the report.

\subsubsection{Training Configurations} We run all experiments on a single machine with Intel(R) Core(TM) i9-10980XE 3.00GHz CPUs, NVIDIA RTX A6000, and $48$ GB RAM memory. We implement UniDyG using PyTorch on CUDA $11.3$. We run $16$ comparative baselines using their official codes. We set the batch size $B$ as $600$. Following the baselines, we also set the neighbor size $N$ as $16$. We train UniDyG with Adam optimizer~\cite{kingma2014adam}, with an empirical learning rate of $0.0001$. 
% Note that all experiments are conducted on a single GPU. 

\begin{table*}[t]
	\caption{Overall performance comparison on four DTDG datasets (\% is omitted). The best results and second best are highlighted in blue and light blue, respectively.}
	\label{tab:3}
	\centering
 \renewcommand{\arraystretch}{1.2}
 \resizebox{\linewidth}{!}{
	\begin{tabular}{|c|c|ccccc|cccc|}
     \Xhline{1.5pt}
		\rowcolor{gray!40}Metric& Dataset & EvolveGCN-H&EvolveGCN-O&Roland &HGWaveNet &WinGNN &DGNN-GRU &DGNN-LSTM &DGNN-TF &UniDyG \\
  \hline
  \hline 
  \multirow{4}{*}{\centering AUC ($\uparrow$)} 
  &BA &63.71$\pm$1.1 &68.93$\pm$0.9&90.21$\pm$1.2&\underline{92.62$\pm$0.1} &91.43$\pm$0.3 &91.98$\pm$0.2&91.87$\pm$0.4&92.38$\pm$0.4 &\textbf{92.86$\pm$0.3}\\
&BO & 55.38$\pm$1.5&59.82$\pm$2.2&90.07$\pm$1.3 &\underline{93.84$\pm$0.2} &91.64$\pm$0.7 &89.87$\pm$0.3&91.14$\pm$0.4&91.35$\pm$0.3&\textbf{96.96$\pm$0.4}\\
&UCI &71.99$\pm$1.8 &62.05$\pm$3.0&91.81$\pm$0.4&\underline{94.80$\pm$0.4} &94.05$\pm$0.5
 &87.79$\pm$0.1&87.42$\pm$0.4&88.03$\pm$0.1&\textbf{98.75$\pm$0.2} \\
&SO &- &-&74.56$\pm$8.1&-&\textbf{99.63$\pm$0.0} &-&-&-&\underline{99.56$\pm$0.0}\\
\hline
\multirow{4}{*}{\centering MRR ($\uparrow$)} 
&BA &3.28$\pm$1.0  &2.52$\pm$0.2 &14.52$\pm$0.7 &31.61$\pm$0.2 &\underline{36.74$\pm$4.0 }  &32.89$\pm$0.1&34.05$\pm$0.2 &31.73$\pm$0.2 &\textbf{42.71$\pm$1.2} \\
&BO & 11.27$\pm$1.5& 11.44$\pm$0.5 & 16.54$\pm$1.1 &\underline{41.10$\pm$0.3}  &37.94$\pm$1.6 &34.05$\pm$0.2  &29.85$\pm$0.5 &31.10$\pm$0.2 &\textbf{53.82$\pm$0.9}\\
&UCI &8.17$\pm$1.2& 10.81$\pm$0.6 & 17.84$\pm$0.3& \underline{32.25$\pm$0.2} &21.69$\pm$0.4  & 31.73$\pm$0.3  &20.24$\pm$0.3&23.14$\pm$0.2 & \textbf{63.08$\pm$1.0}\\
&SO &-& - & 27.55$\pm$7.0 &- &\underline{32.51$\pm$1.6} &-&-&-& \textbf{40.16$\pm$0.5}\\
\hline
\end{tabular}}
\end{table*}

% \begin{table*}[t]
% 	\caption{Link Prediction over CTDGs.}
% 	\label{tab:2}
% 	\centering
% 	\begin{tabular}{|c|cc|cc|cc|cc|cc|cc|}
% 		\hline
%      &\multicolumn{12}{c|}{Transductive Setting}\\
%      \hline
% 		\multirow{2}*{Method} & \multicolumn{2}{c|}{Wikipedia} & \multicolumn{2}{c|}{MOOC} & \multicolumn{2}{c|}{AskUbuntu} & \multicolumn{2}{c|}{LastFM} & \multicolumn{2}{c|}{SuperUser}& \multicolumn{2}{c|}{WIKITALK}\\
% 		\cline{2-13}
% 		{} & AP & AUC & AP & AUC & AP & AUC & AP & AUC & AP & AUC& AP & AUC \\
% 		\hline
%         TGN & - & - & - & - & - & - & - & - & - & -& - & -\\
%         Zebra & - & - & - & - & - & - & - & - & - & -& - & -\\
%         Orca & - & - & - & - & - & - & - & - & - & -& - & -\\
%         GraphMixer & - & - & - & - & - & - & - & - & - & -& - & -\\
%         DyGFormer & - & - & - & - & - & - & - & - & - & -& - & -\\
%         TimeSGN & - & - & - & - & - & - & - & - & - & -& - & -\\
%         ETC & - & - & - & - & - & - & - & - & - & -& - & -\\
%         FreeDyG & - & - & - & - & - & - & - & - & - & -& - & -\\
% 		UniDyG & - & - & - & - & - & - & - & - & - & -& - & -\\
% 		\hline
%     \end{tabular}
% \end{table*}

\subsection{Effectiveness Evaluation}
We evaluate the quality of the temporal embeddings on CTDGs and DTDGs and report results in Tables~\ref{tab:2} and~\ref{tab:3}.

\noindent\textbf{Exp 1: Unified Model Comparison.} Overall, our UniDyG is consistently and significantly better than DGNN across all cases under multiple metrics, achieving an average of $23.5\%$ improvement over CTDG scenarios and an average of $24.7\%$ improvement over DTDG scenarios. The results show the effectiveness of our end-to-end modeling and the significance of long-term temporal and structural dependency learning for the unified model. Furthermore, our UniDyG can scale to large dynamic graphs (\textit{e.g.}, WIKITALK and StackOverFlow), but DGNN cannot finish the learning process within $3$ days in our experimental environment. This suggests the good scalability of our UniDyG.

\noindent\textbf{Exp 2: Effectiveness against CTDG-Specific Methods.} (1) Compared to CTDG-specific approaches, UniDyG outperforms eight baselines in all cases with up to $9.5\%$ improvement compared to the second best, validating the significance of our modeling in the frequency domain. (2) Our UniDyG beats the frequency-enhanced FreeDyG in both transductive and inductive learning scenarios since FreeDyG fails to capture long-term dependencies and global structural correlations in the frequency domain. (3) Although co-neighbor encoding techniques are explored to help learn underlying graph patterns in the time domain in DyGFormer and CNEN, our UniDyG is superior to them by a large margin in inductive learning scenarios. This validates the advantage of our FGAT in capturing underlying graph patterns by exploring local and global structural correlations in the frequency domain. (4) Compared to TimeSGN with the divided temporal message passing paradigm, the results show the effectiveness of 
our Fourier graph attention for temporal structure learning and global view of our temporal dynamic model, especially in the inductive setting with an average of $7.2\%$ improvements.

\noindent\textbf{Exp 3: Effectiveness against DTDG-Specific Methods.} (1) Our UniDyG outperforms the best DTDG-specific baselines in almost all cases, achieving an average improvement of $1.9\%$ in AUC and $26.3\%$ in MRR. This suggests that our UniDyG can make more accurate predictions for future snapshots in the DTDG setting owing to our long-term temporal learning. (2) Compared to gradient-based learning (\textit{i.e.}, Roland and WinGNN) for temporal dynamic learning, our UniDyG yields superior performance since these methods emphasize the temporal dependencies between neighboring snapshots while our UniDyG considers global view for both temporal dependencies and structure evolution. (3) Our UniDyG yields better performance than the HGWaveNet, where HGWaveNet models dynamic graphs in hyperbolic space with more volume for hierarchical structures. The advantages of our UniDyG are especially significant in the ranking setting with up to $48.9\%$ improvements, validating the advantage of frequency-domain modeling in exploring underlying graph evolution.

In summary, our frequency-domain approach, UniDyG, demonstrates superior effectiveness and strong scalability compared to three research lines and fifteen time-domain models, offering valuable insights for future exploration of frequency-domain techniques in dynamic graph learning.

\begin{figure*}[t]
    \centering
    \subfloat[AskUbuntu]{\includegraphics[width=0.25\linewidth]{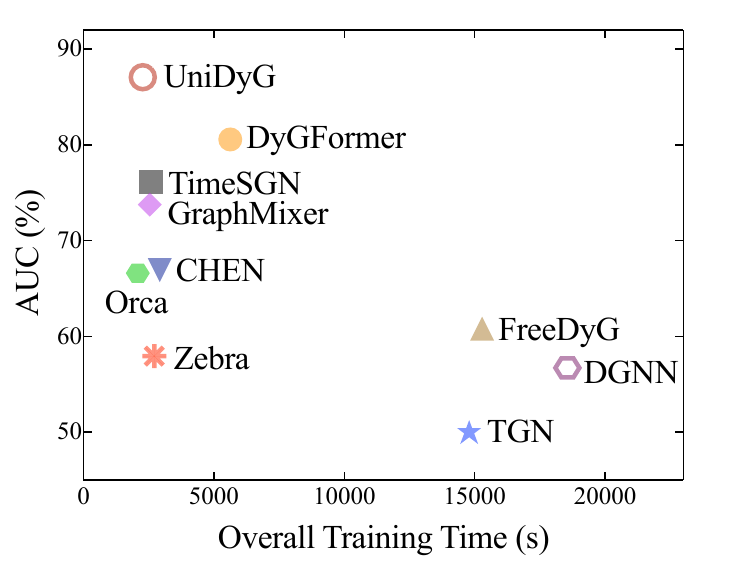}}
    \subfloat[SuperUser]{\includegraphics[width=0.25\linewidth]{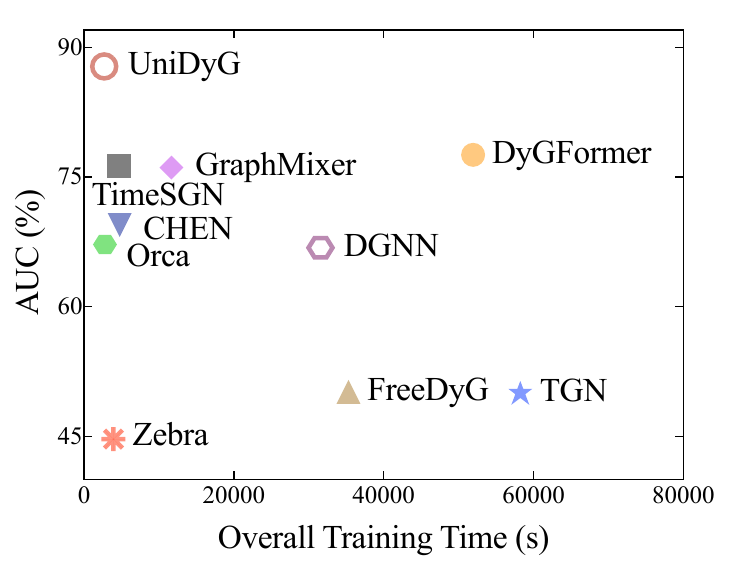}}
    \subfloat[BitcoinOTC]{\includegraphics[width=0.25\linewidth]{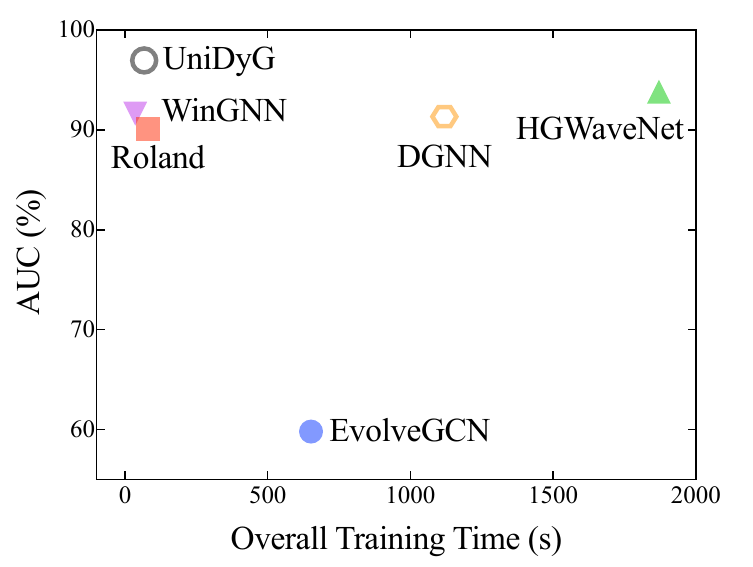}}
    \subfloat[UCI]{\includegraphics[width=0.25\linewidth]{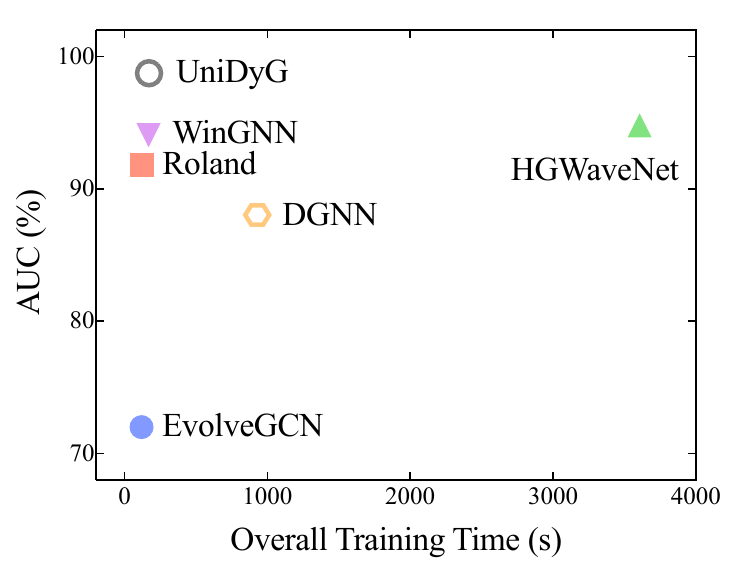}}
    \caption{ Performance Comparison between our UniDyG and baselines on two CTDGs and two DTDGs. The horizontal axis shows the overall training time. The vertical axis shows AUC results.}
    \label{fig:3}
\end{figure*}

\begin{figure*}
    \centering
    \subfloat[AskUbuntu]{\includegraphics[width=0.25\linewidth]{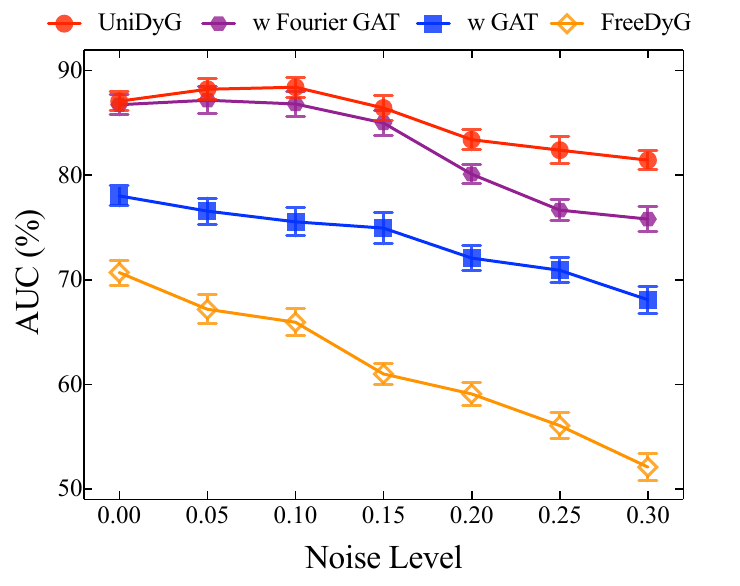}}
    \subfloat[Superuser]{\includegraphics[width=0.25\linewidth]{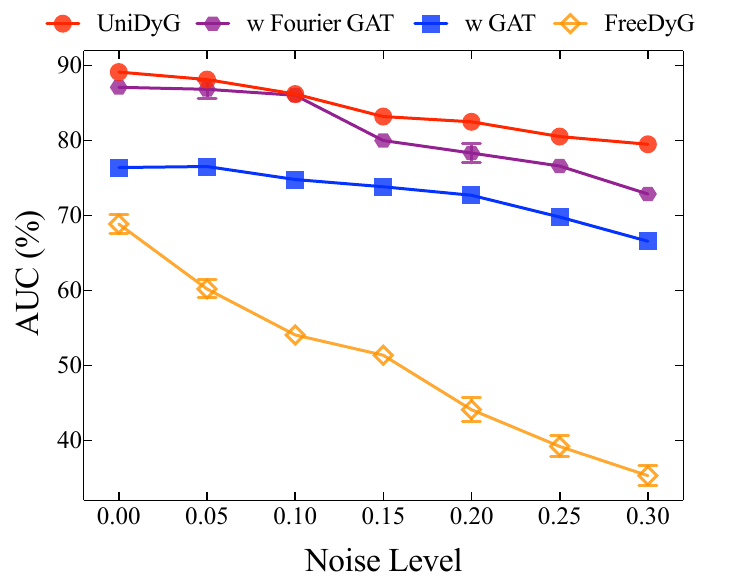}}
    \subfloat[BitcoinOTC]{\includegraphics[width=0.25\linewidth]{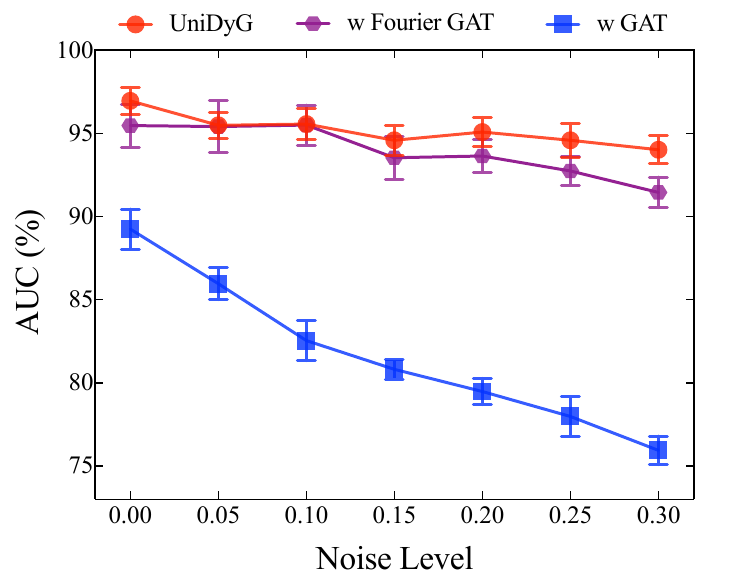}}
    \subfloat[UCI]{\includegraphics[width=0.25\linewidth]{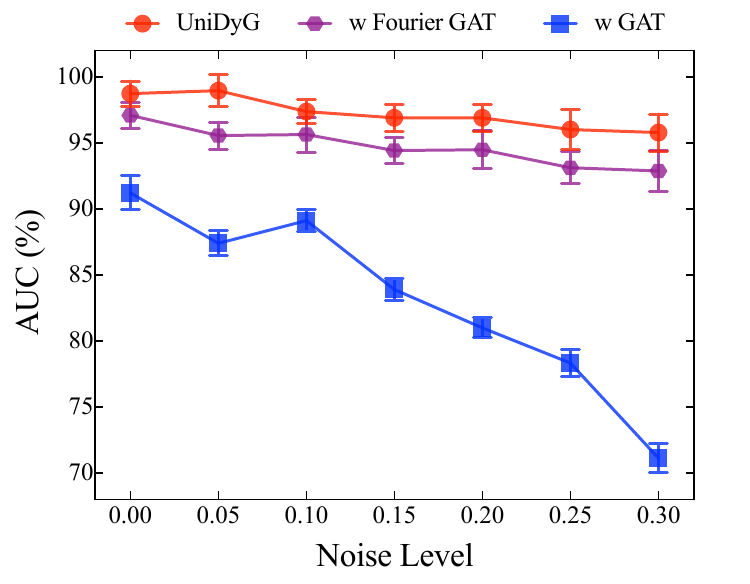}}
    \caption{Robustness evaluation with different level noise in terms of AUC over CTDGs and DTDGs.}
    \label{fig:4}
\end{figure*}

\subsection{Efficiency Evaluation}
To evaluate the efficiency of UniDyG in dynamic graph learning, we compare its overall training time against baselines across four datasets (two CTDGs and two DTDGs). 

\noindent\textbf{Exp 4: Efficiency against Three Groups.} As shown in Fig.~\ref{fig:3}, UniDyG achieves a $5\sim 16\times$ speed-up compared to the unified model DGNN, highlighting the effectiveness of our end-to-end training pipeline. Compared to CTDG-specific methods, UniDyG performs on par with acceleration-driven methods (Zebra, Orca, and TimeSGN) while significantly outperforming them in terms of AUC metric. Besides, UniDyG is faster than FreeDyG by up to $13\times$ on the SuperUser dataset with $1.4$ million edges, due to FreeDyG's more complex time-domain encoding of node interaction frequency. In the DTDG learning scenarios, UniDyG is slightly slower than gradient-based, temporal encoder-free methods such as WinGNN, which benefits from fast convergence. However, UniDyG outperforms them in terms of the AUC metric due to its ability to model global temporal dependencies. Furthermore, UniDyG is better than HGWaveNet in both AUC and overall training time, as HGWaveNet models complex structures in hyperbolic spaces.

\subsection{Robustness Evaluation}
We assess the robustness of UniDyG by evaluating its ability to handle noise across four datasets in terms of AUC. To simulate noise scenarios, we manually introduce varying levels of noise to the dynamic graphs, including noisy edges and noisy attributes. We construct two model variants for comparison: (1) replacing FGAT\_N in attribute-oriented structure learning with FGAT, referred to as `w Fourier GAT'; and (2) using time-domain graph attention for temporal structure learning, referred to as `w GAT'. For CTDG scenarios, we also include FreeDyG in the robustness evaluation. 

\noindent\textbf{Exp 5: Robustness under Noise Scenarios.} As shown in Fig.~\ref{fig:4}, we make the following observations: (1) UniDyG with FGAT\_N demonstrates superior robustness compared to the baselines, as it explicitly filters out high-frequency noise based on energy. (2) At low noise levels, `w Fourier GAT' performs comparably to UniDyG since FGAT is able to capture smooth temporal structure changes through low-frequency bands, enhancing robustness. However, as noise levels increase, the performance gap between UniDyG and `w Fourier GAT' widens, highlighting the effectiveness of FGAT\_N in mitigating noise. (3) The performance of time-domain GAT (`w GAT') drops significantly in both CTDG and DTDG scenarios. This is because time-domain methods tend to overfit noisy information, making them highly sensitive to noise. (4) Despite FreeDyG's learnable filtering in the frequency domain, its robustness is significantly inferior as it lacks specificity in filtering out temporal noise in the frequency domain. Then it treats noisy edges as regular ones and incorporates them into node interaction frequency encoding in the time domain, thus leading to relatively poor robustness.

\begin{table*}[t]
	\caption{Ablation study on nine datasets across CTDG and DTDG settings.}
	\label{tab:4}
	\centering
 \renewcommand{\arraystretch}{1.2}
 \resizebox{\linewidth}{!}{
	\begin{tabular}{|c|cccc|ccccc|}
     \Xhline{1.5pt}
     \rowcolor{gray!40} Method& BitcoinAlpha &BitcoinOTC &UCI &Stack Overflow&Wikipedia &AskUbuntu &LastFM &SuperUser &WIKITALK \\ 
     \hline
     \hline 
    Type&\multicolumn{4}{>{\columncolor{gray!5}}c|}{DTDG (AUC)}& \multicolumn{5}{>{\columncolor{gray!20}}c|}{CTDG (AUC, Inductive)}\\
  \hline
UniDyG & 92.86& 96.96& 98.75 & 99.56 & 99.29& 87.09& 99.89& 89.16&86.75\\
`w/o FGAT\_N' &90.01 &95.48&97.09&98.77&99.14&86.76&98.63&87.11&80.82\\
`w/o Global' &84.70&93.53&96.01&97.99&98.96&86.70&95.05&87.58 &84.67\\
`w GAT' &85.61 &89.25&91.23&93.47&98.69&76.58&96.26&76.40&74.78\\
% `w/o Dual' &89.64 &96.04&97.94 &98.12 &98.65 &86.12 &99.19 &85.52 & 81.41\\
\hline
\end{tabular}}
\end{table*}

\subsection{Investigation of UniDyG}
\noindent\textbf{Exp 6: Ablation Study.}
To evaluate the effectiveness of each component in UniDyG, we compare the full model with its three variants across nine datasets over CTDG and DTDG settings. The constructed variants include (1) `w/o FGAT\_N': replacing FGAT\_N with FGAT; (2) `w/o Global': substituting the frequency-enhanced linear function with a time-domain linear function; and (3) `w GAT': using time-domain GAT for temporal structure learning. The results in Table~\ref{tab:4} reveal several key observations. (1) FGAT\_N proves to be particularly crucial in inductive learning scenarios on CTDG datasets, achieving up to a $6.8\% $ improvement. This is because FGAT\_N not only filters out noisy information but also prevents overfitting to observed data, thereby enhancing model generalization in the inductive setting. (2) A global view of temporal dependencies is essential for DTDG learning within our training pipeline, as it helps capture dependencies across different snapshots. In contrast, a simple time-domain linear function fails to capture long-term temporal dependencies between snapshots, leading to reduced model performance. (3) In both CTDG and DTDG learning scenarios, time-domain graph attention significantly reduces performance in terms of AUC by an average of $8.2\%$, validating the superior abilities of our Fourier graph attention in capturing complex temporal structures across various dynamic graph types.

\begin{figure}[t]
    \centering
    \subfloat[Askubuntu]{\includegraphics[width=0.5\linewidth]{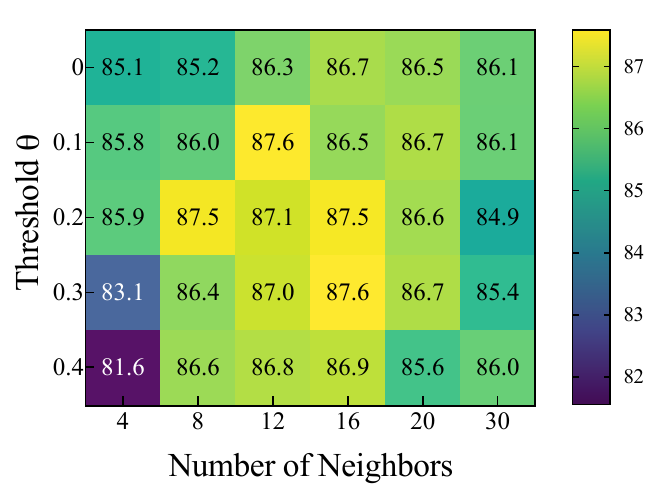}}
    \subfloat[BitcoinOTC]{\includegraphics[width=0.5\linewidth]{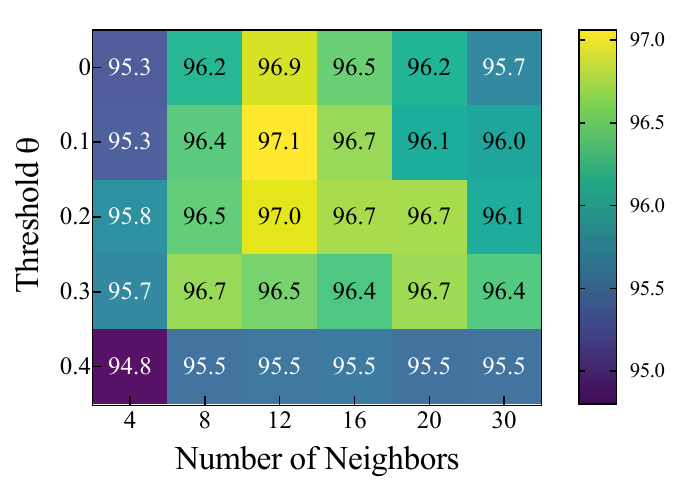}}
    \caption{The heatmap values in AUC (\%) of UniDyG across different parameter settings.}
    \label{fig:5}
\end{figure}

\noindent\textbf{Exp 7: Parameter Sensitivity.} We investigate the impact of two key parameters in UniDyG: the number of neighbors $N$ and the threshold $\theta$ of the energy-gated unit. We report the results over two datasets in terms of the AUC metric in Fig.~\ref{fig:5}. We observe that a relatively smaller $\theta$ leads to better performance, as it effectively filters out high-frequency temporal noise while preserving valuable abrupt changes. In contrast, a larger $\theta$ (\textit{e.g.}, $0.4$) significantly compromises model performance by filtering out too much valuable information. Furthermore, UniDyG demonstrates robustness across a wide range of neighbor counts (\textit{i.e.}, $8\sim 20$). Meanwhile, the smaller or larger number of neighbors would incur underfitting or overfitting issues, respectively, thus compromising model performance in the inductive setting. Throughout this paper, we set $\theta$ and $N$ as $0.2$ and $12$.

% \section{Discussion}
% In this section, we further analyze HTGN with the aim of answering
% the following research questions:
% \begin{itemize}
%     \item What does hyperbolic geometry bring?
%     \item How is the learning efficiency in large networks?
% \end{itemize}

\section{Conclusion}
In this paper, we present UniDyG, a unified representation learning approach for both continuous-time and discrete-time dynamic graphs, capable of handling large dynamic graphs. To model different temporal structure evolutions, we propose a novel Fourier graph attention mechanism and an optimized variant that can learn local and global structural correlations using recent neighbors and complex-number selective aggregation, which can generate high-quality temporal representations while addressing the noise issue. We provide the theoretical guarantee from views of temporal coherence and universal approximation. Upon our FGAT mechanisms, our UniDyG can model long-term temporal and structural dependencies with a unified and end-to-end training pipeline, making it adaptable for both CTDGs and DTDGs. Extensive results validate its effectiveness, scalability, and robustness.

This paper provides valuable insights for future work in frequency-domain representation learning for dynamic graphs. Exploring frequency-domain techniques offers promising opportunities for capturing long-term temporal and structural evolutions underlying dynamic graphs with good scalability.

% \section*{Acknowledgments}
% This should be a simple paragraph before the References to thank those individuals and institutions who have supported your work on this article.

% {\appendix[Proof of the Zonklar Equations]
% Use $\backslash${\tt{appendix}} if you have a single appendix:
% Do not use $\backslash${\tt{section}} anymore after $\backslash${\tt{appendix}}, only $\backslash${\tt{section*}}.
% If you have multiple appendixes use $\backslash${\tt{appendices}} then use $\backslash${\tt{section}} to start each appendix.
% You must declare a $\backslash${\tt{section}} before using any $\backslash${\tt{subsection}} or using $\backslash${\tt{label}} ($\backslash${\tt{appendices}} by itself
%  starts a section numbered zero.)}



\bibliographystyle{IEEEtran}
\bibliography{reference}

\begin{thebibliography}{10}
\providecommand{\url}[1]{#1}
\csname url@samestyle\endcsname
\providecommand{\newblock}{\relax}
\providecommand{\bibinfo}[2]{#2}
\providecommand{\BIBentrySTDinterwordspacing}{\spaceskip=0pt\relax}
\providecommand{\BIBentryALTinterwordstretchfactor}{4}
\providecommand{\BIBentryALTinterwordspacing}{\spaceskip=\fontdimen2\font plus
\BIBentryALTinterwordstretchfactor\fontdimen3\font minus \fontdimen4\font\relax}
\providecommand{\BIBforeignlanguage}[2]{{%
\expandafter\ifx\csname l@#1\endcsname\relax
\typeout{** WARNING: IEEEtran.bst: No hyphenation pattern has been}%
\typeout{** loaded for the language `#1'. Using the pattern for}%
\typeout{** the default language instead.}%
\else
\language=\csname l@#1\endcsname
\fi
#2}}
\providecommand{\BIBdecl}{\relax}
\BIBdecl

\bibitem{sun2022aligning}
L.~Sun, Z.~Zhang, F.~Wang, P.~Ji, J.~Wen, S.~Su, and S.~Y. Philip, ``Aligning dynamic social networks: An optimization over dynamic graph autoencoder,'' \emph{IEEE Transactions on Knowledge and Data Engineering}, vol.~35, no.~6, pp. 5597--5611, 2022.

\bibitem{liu2021anomaly}
Y.~Liu, S.~Pan, Y.~G. Wang, F.~Xiong, L.~Wang, Q.~Chen, and V.~C. Lee, ``Anomaly detection in dynamic graphs via transformer,'' \emph{IEEE Transactions on Knowledge and Data Engineering}, vol.~35, no.~12, pp. 12\,081--12\,094, 2021.

\bibitem{DBLP:journals/jmlr/KazemiGJKSFP20}
S.~M. Kazemi, R.~Goel, K.~Jain, I.~Kobyzev, A.~Sethi, P.~Forsyth, and P.~Poupart, ``Representation learning for dynamic graphs: {A} survey,'' \emph{J. Mach. Learn. Res.}, vol.~21, pp. 70:1--70:73, 2020.

\bibitem{DBLP:conf/bigdataconf/FuH22}
D.~Fu and J.~He, ``{DPPIN:} {A} biological repository of dynamic protein-protein interaction network data,'' in \emph{Proceedings of The {IEEE} International Conference on Big Data}.\hskip 1em plus 0.5em minus 0.4em\relax {IEEE}, 2022, pp. 5269--5277.

\bibitem{DBLP:conf/kdd/KumarZL19}
S.~Kumar, X.~Zhang, and J.~Leskovec, ``Predicting dynamic embedding trajectory in temporal interaction networks,'' in \emph{Proceedings of The {ACM} {SIGKDD} International Conference on Knowledge Discovery {\&} Data Mining}.\hskip 1em plus 0.5em minus 0.4em\relax {ACM}, 2019, pp. 1269--1278.

\bibitem{DBLP:conf/iclr/TrivediFBZ19}
R.~Trivedi, M.~Farajtabar, P.~Biswal, and H.~Zha, ``Dyrep: Learning representations over dynamic graphs,'' in \emph{Proceedings of The International Conference on Learning Representations ({ICLR})}.\hskip 1em plus 0.5em minus 0.4em\relax OpenReview.net, 2019.

\bibitem{tgn_icml_grl2020}
E.~Rossi, B.~Chamberlain, F.~Frasca, D.~Eynard, F.~Monti, and M.~Bronstein, ``Temporal graph networks for deep learning on dynamic graphs,'' in \emph{ICML 2020 Workshop on Graph Representation Learning}, 2020.

\bibitem{zhou2022tgl}
H.~Zhou, D.~Zheng, I.~Nisa, V.~Ioannidis, X.~Song, and G.~Karypis, ``Tgl: a general framework for temporal gnn training on billion-scale graphs,'' \emph{Proceedings of The VLDB Endowment}, vol.~15, no.~8, pp. 1572--1580, 2022.

\bibitem{DBLP:journals/pacmmod/LiSCY23}
Y.~Li, Y.~Shen, L.~Chen, and M.~Yuan, ``Orca: Scalable temporal graph neural network training with theoretical guarantees,'' \emph{Proceedings of The International Conference on Management of Data (SIGMOD)}, vol.~1, no.~1, pp. 52:1--52:27, 2023.

\bibitem{gao2024simple}
S.~Gao, Y.~Li, X.~Zhang, Y.~Shen, Y.~Shao, and L.~Chen, ``Simple: Efficient temporal graph neural network training at scale with dynamic data placement,'' \emph{Proceedings of the ACM on Management of Data}, vol.~2, no.~3, pp. 1--25, 2024.

\bibitem{cheng2024co}
K.~Cheng, P.~Linzhi, J.~Ye, L.~Sun, and B.~Du, ``Co-neighbor encoding schema: A light-cost structure encoding method for dynamic link prediction,'' in \emph{Proceedings of the 30th ACM SIGKDD Conference on Knowledge Discovery and Data Mining}, 2024, pp. 421--432.

\bibitem{TimeSGN}
Y.~Xu, W.~Zhang, Y.~Zhang, M.~Orlowska, and X.~Lin, ``Timesgn: Scalable and effective temporal graph neural network,'' in \emph{Proceedings of IEEE International Conference on Data Engineering (ICDE)}, 2024, pp. 3297--3310.

\bibitem{DBLP:conf/icdm/SharanN08}
U.~Sharan and J.~Neville, ``Temporal-relational classifiers for prediction in evolving domains,'' in \emph{Proceedings of The {IEEE} International Conference on Data Mining (ICDM)}.\hskip 1em plus 0.5em minus 0.4em\relax {IEEE} Computer Society, 2008, pp. 540--549.

\bibitem{DBLP:conf/www/ZhangYJL23}
G.~Zhang, T.~Ye, D.~Jin, and Y.~Li, ``An attentional multi-scale co-evolving model for dynamic link prediction,'' in \emph{Proceedings of The {ACM} Web Conference}.\hskip 1em plus 0.5em minus 0.4em\relax {ACM}, 2023, pp. 429--437.

\bibitem{DBLP:conf/aaai/LiYZC0ZTWM23}
J.~Li, Z.~Yu, Z.~Zhu, L.~Chen, Q.~Yu, Z.~Zheng, S.~Tian, R.~Wu, and C.~Meng, ``Scaling up dynamic graph representation learning via spiking neural networks,'' in \emph{Proceedings of The {AAAI} Conference on Artificial Intelligence}.\hskip 1em plus 0.5em minus 0.4em\relax {AAAI} Press, 2023, pp. 8588--8596.

\bibitem{zhu2023wingnn}
Y.~Zhu, F.~Cong, D.~Zhang, W.~Gong, Q.~Lin, W.~Feng, Y.~Dong, and J.~Tang, ``Wingnn: Dynamic graph neural networks with random gradient aggregation window,'' in \emph{Proceedings of the ACM SIGKDD Conference on Knowledge Discovery and Data Mining}, 2023, pp. 3650--3662.

\bibitem{DBLP:conf/kdd/YouDL22}
J.~You, T.~Du, and J.~Leskovec, ``{ROLAND:} graph learning framework for dynamic graphs,'' in \emph{Proceedings of The {ACM} {SIGKDD} Conference on Knowledge Discovery and Data Mining}, A.~Zhang and H.~Rangwala, Eds.\hskip 1em plus 0.5em minus 0.4em\relax {ACM}, 2022, pp. 2358--2366.

\bibitem{hao2023dynamic}
Y.~Hao, Y.~Mao, X.~Cao, Y.~Fang, X.~Lin, and H.~Mao, ``Dynamic graph embedding via meta-learning,'' \emph{IEEE Transactions on Knowledge and Data Engineering}, 2023.

\bibitem{zhang2023dyted}
K.~Zhang, Q.~Cao, G.~Fang, B.~Xu, H.~Zou, H.~Shen, and X.~Cheng, ``Dyted: Disentangled representation learning for discrete-time dynamic graph,'' in \emph{Proceedings of the ACM SIGKDD Conference on Knowledge Discovery and Data Mining}, 2023, pp. 3309--3320.

\bibitem{zhao2024adversarial}
Z.~Zhao, Y.~Yang, Z.~Yin, T.~Xu, X.~Zhu, F.~Lin, X.~Li, and E.~Chen, ``Adversarial attack and defense on discrete time dynamic graphs,'' \emph{IEEE Transactions on Knowledge and Data Engineering}, 2024.

\bibitem{DBLP:journals/pvldb/ZhengWL23}
Y.~Zheng, Z.~Wei, and J.~Liu, ``Decoupled graph neural networks for large dynamic graphs,'' \emph{Proceedings of The VLDB Endowment}, vol.~16, no.~9, pp. 2239--2247, 2023.

\bibitem{DBLP:conf/iclr/XuRKKA20}
D.~Xu, C.~Ruan, E.~K{\"{o}}rpeoglu, S.~Kumar, and K.~Achan, ``Inductive representation learning on temporal graphs,'' in \emph{Proceedings of The International Conference on Learning Representations (ICLR)}.\hskip 1em plus 0.5em minus 0.4em\relax OpenReview.net, 2020.

\bibitem{yang2021time}
Y.~Yang, J.~Cao, M.~Stojmenovic, S.~Wang, Y.~Cheng, C.~Lum, and Z.~Li, ``Time-capturing dynamic graph embedding for temporal linkage evolution,'' \emph{IEEE Transactions on Knowledge and Data Engineering}, vol.~35, no.~1, pp. 958--971, 2021.

\bibitem{liu2020k}
J.~Liu, C.~Xu, C.~Yin, W.~Wu, and Y.~Song, ``K-core based temporal graph convolutional network for dynamic graphs,'' \emph{IEEE Transactions on Knowledge and Data Engineering}, vol.~34, no.~8, pp. 3841--3853, 2020.

\bibitem{DBLP:conf/iclr/CongZKYWZTM23}
W.~Cong, S.~Zhang, J.~Kang, B.~Yuan, H.~Wu, X.~Zhou, H.~Tong, and M.~Mahdavi, ``Do we really need complicated model architectures for temporal networks?'' in \emph{Proceedings of International Conference on Learning Representations (ICLR)}.\hskip 1em plus 0.5em minus 0.4em\relax OpenReview.net, 2023.

\bibitem{DBLP:journals/pvldb/FangFGFH23}
L.~Fang, K.~Feng, J.~Gui, S.~Feng, and A.~Hu, ``Anonymous edge representation for inductive anomaly detection in dynamic bipartite graphs,'' \emph{Proceedings of The VLDB Endowment}, vol.~16, no.~5, pp. 1154--1167, 2023.

\bibitem{DBLP:conf/www/SureshSMN023}
S.~Suresh, M.~Shrivastava, A.~Mukherjee, J.~Neville, and P.~Li, ``Expressive and efficient representation learning for ranking links in temporal graphs,'' in \emph{Proceedings of The {ACM} Web Conference}.\hskip 1em plus 0.5em minus 0.4em\relax {ACM}, 2023, pp. 567--577.

\bibitem{xu2024scalable}
Y.~Xu, W.~Zhang, X.~Xu, B.~Li, and Y.~Zhang, ``Scalable and effective temporal graph representation learning with hyperbolic geometry,'' \emph{IEEE Transactions on Neural Networks and Learning Systems}, 2024.

\bibitem{xu2024bootstrapping}
Y.~Xu, L.~Han, L.~Sun, B.~Du, C.~Liu, and H.~Xiong, ``Bootstrapping on continuous-time dynamic graphs for crowd flow modeling,'' \emph{IEEE Transactions on Knowledge and Data Engineering}, 2024.

\bibitem{DBLP:conf/iclr/WangCLL021}
Y.~Wang, Y.~Chang, Y.~Liu, J.~Leskovec, and P.~Li, ``Inductive representation learning in temporal networks via causal anonymous walks,'' in \emph{Proceedings of International Conference on Learning Representations (ICLR)}.\hskip 1em plus 0.5em minus 0.4em\relax OpenReview.net, 2021.

\bibitem{souza2022provably}
A.~Souza, D.~Mesquita, S.~Kaski, and V.~Garg, ``Provably expressive temporal graph networks,'' \emph{Advances in neural information processing systems}, vol.~35, pp. 32\,257--32\,269, 2022.

\bibitem{DBLP:conf/nips/JinLP22}
M.~Jin, Y.~Li, and S.~Pan, ``Neural temporal walks: Motif-aware representation learning on continuous-time dynamic graphs,'' in \emph{Advances in Neural Information Processing Systems (NeurIPS)}, 2022.

\bibitem{luo2022neighborhood}
Y.~Luo and P.~Li, ``Neighborhood-aware scalable temporal network representation learning,'' in \emph{Proceedings of the Learning on Graphs Conference}.\hskip 1em plus 0.5em minus 0.4em\relax PMLR, 2022, pp. 1--1.

\bibitem{yu2023towards}
L.~Yu, L.~Sun, B.~Du, and W.~Lv, ``Towards better dynamic graph learning: New architecture and unified library,'' \emph{Advances in Neural Information Processing Systems}, vol.~36, pp. 67\,686--67\,700, 2023.

\bibitem{tian2023freedyg}
Y.~Tian, Y.~Qi, and F.~Guo, ``Freedyg: Frequency enhanced continuous-time dynamic graph model for link prediction,'' in \emph{Proceedings of The International Conference on Learning Representations}, 2023.

\bibitem{li2024robust}
H.~Li, C.~Li, K.~Feng, Y.~Yuan, G.~Wang, and H.~Zha, ``Robust knowledge adaptation for dynamic graph neural networks,'' \emph{IEEE Transactions on Knowledge and Data Engineering}, 2024.

\bibitem{DBLP:conf/sigmod/WangLLXYWWCYSG21}
X.~Wang, D.~Lyu, M.~Li, Y.~Xia, Q.~Yang, X.~Wang, X.~Wang, P.~Cui, Y.~Yang, B.~Sun, and Z.~Guo, ``{APAN:} asynchronous propagation attention network for real-time temporal graph embedding,'' in \emph{Proceedings of The International Conference on Management of Data (SIGMOD)}.\hskip 1em plus 0.5em minus 0.4em\relax {ACM}, 2021, pp. 2628--2638.

\bibitem{yang2023time}
Y.~Yang, H.~Yin, J.~Cao, T.~Chen, Q.~V.~H. Nguyen, X.~Zhou, and L.~Chen, ``Time-aware dynamic graph embedding for asynchronous structural evolution,'' \emph{IEEE Transactions on Knowledge and Data Engineering}, vol.~35, no.~9, pp. 9656--9670, 2023.

\bibitem{DBLP:journals/pvldb/LiSCY23}
Y.~Li, Y.~Shen, L.~Chen, and M.~Yuan, ``Zebra: When temporal graph neural networks meet temporal personalized pagerank,'' \emph{Proceedings of The VLDB Endowment}, vol.~16, no.~6, pp. 1332--1345, 2023.

\bibitem{li2017attributed}
J.~Li, H.~Dani, X.~Hu, J.~Tang, Y.~Chang, and H.~Liu, ``Attributed network embedding for learning in a dynamic environment,'' in \emph{Proceedings of the ACM on Conference on Information and Knowledge Management}, 2017, pp. 387--396.

\bibitem{DBLP:conf/icde/ZhuGYSG17}
L.~Zhu, D.~Guo, J.~Yin, G.~V. Steeg, and A.~Galstyan, ``Scalable temporal latent space inference for link prediction in dynamic social networks (extended abstract),'' in \emph{Proceedings of The {IEEE} International Conference on Data Engineering ({ICDE)}}.\hskip 1em plus 0.5em minus 0.4em\relax {IEEE} Computer Society, 2017, pp. 57--58.

\bibitem{DBLP:conf/iconip/SeoDVB18}
Y.~Seo, M.~Defferrard, P.~Vandergheynst, and X.~Bresson, ``Structured sequence modeling with graph convolutional recurrent networks,'' in \emph{Proceedings of The International Conference on Neural Information Processing ({ICONIP)}}, ser. Lecture Notes in Computer Science, vol. 11301.\hskip 1em plus 0.5em minus 0.4em\relax Springer, 2018, pp. 362--373.

\bibitem{DBLP:conf/wsdm/SankarWGZY20}
A.~Sankar, Y.~Wu, L.~Gou, W.~Zhang, and H.~Yang, ``Dysat: Deep neural representation learning on dynamic graphs via self-attention networks,'' in \emph{Proceedings of The {ACM} International Conference on Web Search and Data Mining (WSDM)}, J.~Caverlee, X.~B. Hu, M.~Lalmas, and W.~Wang, Eds.\hskip 1em plus 0.5em minus 0.4em\relax {ACM}, 2020, pp. 519--527.

\bibitem{DBLP:conf/aaai/ParejaDCMSKKSL20}
A.~Pareja, G.~Domeniconi, J.~Chen, T.~Ma, T.~Suzumura, H.~Kanezashi, T.~Kaler, T.~B. Schardl, and C.~E. Leiserson, ``Evolvegcn: Evolving graph convolutional networks for dynamic graphs,'' in \emph{Proceedings of The {AAAI} Conference on Artificial Intelligence}.\hskip 1em plus 0.5em minus 0.4em\relax {AAAI} Press, 2020, pp. 5363--5370.

\bibitem{gao2022novel}
C.~Gao, J.~Zhu, F.~Zhang, Z.~Wang, and X.~Li, ``A novel representation learning for dynamic graphs based on graph convolutional networks,'' \emph{IEEE Transactions on Cybernetics}, vol.~53, no.~6, pp. 3599--3612, 2022.

\bibitem{DBLP:conf/www/BaiNZZY23}
Q.~Bai, C.~Nie, H.~Zhang, D.~Zhao, and X.~Yuan, ``Hgwavenet: {A} hyperbolic graph neural network for temporal link prediction,'' in \emph{Proceedings of The {ACM} Web Conference}.\hskip 1em plus 0.5em minus 0.4em\relax {ACM}, 2023, pp. 523--532.

\bibitem{DBLP:conf/icde/0003SLRD23}
X.~Qin, N.~Sheikh, C.~Lei, B.~Reinwald, and G.~Domeniconi, ``{SEIGN:} {A} simple and efficient graph neural network for large dynamic graphs,'' in \emph{Proceedings of The {IEEE} International Conference on Data Engineering (ICDE)}.\hskip 1em plus 0.5em minus 0.4em\relax {IEEE}, 2023, pp. 2850--2863.

\bibitem{DBLP:conf/kdd/0001ZKHK21}
M.~Yang, M.~Zhou, M.~Kalander, Z.~Huang, and I.~King, ``Discrete-time temporal network embedding via implicit hierarchical learning in hyperbolic space,'' in \emph{Proceedings of The {ACM} {SIGKDD} Conference on Knowledge Discovery and Data Mining}.\hskip 1em plus 0.5em minus 0.4em\relax {ACM}, 2021, pp. 1975--1985.

\bibitem{DBLP:conf/cikm/LiC21}
H.~Li and L.~Chen, ``Cache-based {GNN} system for dynamic graphs,'' in \emph{Proceedings of The {ACM} International Conference on Information and Knowledge Management}.\hskip 1em plus 0.5em minus 0.4em\relax {ACM}, 2021, pp. 937--946.

\bibitem{qin2023high}
M.~Qin, C.~Zhang, B.~Bai, G.~Zhang, and D.-Y. Yeung, ``High-quality temporal link prediction for weighted dynamic graphs via inductive embedding aggregation,'' \emph{IEEE Transactions on Knowledge and Data Engineering}, vol.~35, no.~9, pp. 9378--9393, 2023.

\bibitem{zou2023event}
T.~Zou, L.~Yu, L.~Sun, B.~Du, D.~Wang, and F.~Zhuang, ``Event-based dynamic graph representation learning for patent application trend prediction,'' \emph{IEEE Transactions on Knowledge and Data Engineering}, 2023.

\bibitem{kong2023efficient}
L.~Kong, J.~Dong, J.~Ge, M.~Li, and J.~Pan, ``Efficient frequency domain-based transformers for high-quality image deblurring,'' in \emph{Proceedings of the IEEE/CVF Conference on Computer Vision and Pattern Recognition}, 2023, pp. 5886--5895.

\bibitem{zhang2017stock}
L.~Zhang, C.~Aggarwal, and G.-J. Qi, ``Stock price prediction via discovering multi-frequency trading patterns,'' in \emph{Proceedings of the ACM SIGKDD international conference on knowledge discovery and data mining}, 2017, pp. 2141--2149.

\bibitem{yang2020adaptive}
Z.~Yang, W.~Yan, X.~Huang, and L.~Mei, ``Adaptive temporal-frequency network for time-series forecasting,'' \emph{IEEE Transactions on Knowledge and Data Engineering}, vol.~34, no.~4, pp. 1576--1587, 2020.

\bibitem{zhou2022fedformer}
T.~Zhou, Z.~Ma, Q.~Wen, X.~Wang, L.~Sun, and R.~Jin, ``Fedformer: Frequency enhanced decomposed transformer for long-term series forecasting,'' in \emph{Proceedings of International conference on machine learning}.\hskip 1em plus 0.5em minus 0.4em\relax PMLR, 2022, pp. 27\,268--27\,286.

\bibitem{DBLP:conf/icml/Eldele0C0024}
E.~Eldele, M.~Ragab, Z.~Chen, M.~Wu, and X.~Li, ``Tslanet: Rethinking transformers for time series representation learning,'' in \emph{Proceedings of the International Conference on Machine Learning}, 2024.

\bibitem{DBLP:journals/debu/HamiltonYL17}
W.~L. Hamilton, R.~Ying, and J.~Leskovec, ``Representation learning on graphs: Methods and applications,'' \emph{{IEEE} Data Eng. Bull.}, vol.~40, no.~3, pp. 52--74, 2017.

\bibitem{yi2024frequency}
K.~Yi, Q.~Zhang, W.~Fan, S.~Wang, P.~Wang, H.~He, N.~An, D.~Lian, L.~Cao, and Z.~Niu, ``Frequency-domain mlps are more effective learners in time series forecasting,'' \emph{Advances in Neural Information Processing Systems}, vol.~36, 2024.

\bibitem{brigham1988fast}
E.~Brigham, ``The fast fourier transform and its applications,'' 1988.

\bibitem{stone1948generalized}
M.~H. Stone, ``The generalized weierstrass approximation theorem,'' \emph{Mathematics Magazine}, vol.~21, no.~5, pp. 237--254, 1948.

\bibitem{paranjape2017motifs}
A.~Paranjape, A.~R. Benson, and J.~Leskovec, ``Motifs in temporal networks,'' in \emph{Proceedings of the tenth ACM international conference on web search and data mining}, 2017, pp. 601--610.

\bibitem{pareja2020evolvegcn}
A.~Pareja, G.~Domeniconi, J.~Chen, T.~Ma, T.~Suzumura, H.~Kanezashi, T.~Kaler, T.~Schardl, and C.~Leiserson, ``Evolvegcn: Evolving graph convolutional networks for dynamic graphs,'' in \emph{Proceedings of the AAAI conference on artificial intelligence}, vol.~34, no.~04, 2020, pp. 5363--5370.

\bibitem{kingma2014adam}
D.~P. Kingma and J.~Ba, ``Adam: A method for stochastic optimization,'' \emph{arXiv preprint arXiv:1412.6980}, 2014.

\end{thebibliography}

\vfill

\end{document}